\documentclass[stsy]{informs-stsy}

\DoubleSpacedXI

\usepackage{custom}

\usepackage{algorithm}
\usepackage{algorithmic}
\usepackage[caption=false]{subfig}
\newcommand\norm[1]{\lVert#1\rVert}

\usepackage{multirow}
\usepackage{amsmath}

\usepackage{amssymb}
\usepackage{enumitem}
\usepackage{algorithm}

\usepackage{natbib}
 \bibpunct[, ]{(}{)}{,}{a}{}{,}%
 \def\bibfont{\small}%
\TheoremsNumberedThrough     
\ECRepeatTheorems

\EquationsNumberedThrough    

\MANUSCRIPTNO{MS-0001-1922.65}

\begin{document}


 \RUNAUTHOR{Winnicki and Srikant}

\RUNTITLE{A New Policy Iteration Algorithm For Zero-Sum Markov Games}

\TITLE{A New Policy Iteration Algorithm For Reinforcement Learning In Zero-Sum Markov Games}

\ARTICLEAUTHORS{%
\AUTHOR{ Anna Winnicki  }
\AFF{University of Illinois  Urbana-Champaign, Coordinated Science Laboratory and Department of Electrical and Computer Engineering, Urbana, IL, 61801, annaw5@illinois.edu} 
\AUTHOR{R. Srikant}
\AFF{University of Illinois  Urbana-Champaign, Coordinated Science Laboratory and Department of Electrical and Computer Engineering, Urbana, IL, 61801, \EMAIL{rsrikant@illinois.edu}. R. Srikant is also affiliated with c3.ai DTI.}
} 

\ABSTRACT{
Optimal policies in standard MDPs can be obtained using either value iteration or policy iteration. However, in the case of zero-sum Markov games, there is no efficient policy iteration algorithm; e.g., it has been shown that one has to solve $\Omega(1/(1-\alpha))$ MDPs, where $\alpha$ is the discount factor, to implement the only known convergent version of policy iteration. Another algorithm, called naive policy iteration, is easy to implement but is only provably convergent under very restrictive assumptions. Prior attempts to fix naive policy iteration algorithm have several limitations. Here, we show that a simple variant of naive policy iteration for games converges exponentially fast. The only addition we propose to naive policy iteration is the use of lookahead policies, which are anyway used in practical algorithms. We further show that lookahead can be implemented efficiently in the function approximation setting of linear Markov games, which are the counterpart of the much-studied linear MDPs. We illustrate the application of our algorithm by providing bounds for policy-based RL (reinforcement learning) algorithms. We extend the results to the function approximation setting.
}



\maketitle
\section{Introduction}

Multi-agent reinforcement learning algorithms have contributed to many successes in machine learning, including games such as chess and Go \cite{silver2016mastering, DBLP:journals/corr/MnihBMGLHSK16, silver2017mastering, silver2017shoji}, automated warehouses, robotic arms with multiple arms \cite{gu2017deep}, and autonomous traffic control \cite{yang2020multi, shalev2016safe}; see \cite{ozdaglar2021independent, zhang2021multi, yang2020overview} for surveys. Multi-agent RL can refer to one of many scenarios: (i) where a team of agents work towards a common goal where all the agents have the same information or not \cite{qu2022scalable}, (ii) non-cooperative games where there are multiple agents with their own objectives \cite{zhang2019non}, and (iii) zero-sum games where there are only two players with opposing objectives. We can further categorize problems as infinite-horizon, discounted reward/cost, finite-horizon, simultaneous move or turn-based games. The literature in this area is vast; here we focus on zero-sum, simultaneous move, discounted reward/cost Markov games.

In the case of model-based zero-sum discounted reward/cost simultaneous-move Markov games, the problem of interest is finding a Nash equilibrium strategy. In the setting of Markov Games (also known as Stochastic Games \cite{Shapley}), the problem of finding the Nash equilibrium is a generalization of finding an optimal policy for a Markov Decision process \cite{Puterman1978ModifiedPI, lagoudakis2012value}, however many algorithms that find optimal policies for MDPs cannot efficiently be extended to the Markov Games setting, largely due to monotonicity issues that arise from the competing objectives of the two players as opposed to the single entity setting of MDPs. 

While value iteration naturally extends to zero-sum Markov games \cite{Shapley}, 
the two main extensions of Howard's Policy Iteration (PI) \cite{Puterman1978ModifiedPI} for games are not very computationally efficient or simply do not converge.  The only known convergent algorithm, the Hoffman and Karp algorithm \cite{hoffman1966nonterminating}, requires solving an MDP at each iteration. 
It has been shown in \cite{hansen2013strategy} that one has to solve $\Omega(1/(1-\alpha))$ MDPs to implement the extension of PI for games. 

There is an alternative algorithm, called naive policy iteration, also known as the algorithm of Pollatschek and Avi-Itzhak, which requires far fewer computations, but is only known to converge that under restrictive assumptions \cite{pollatschek1969algorithms}. In fact, \cite{van1978discounted} shows that the algorithm does not converge, in general.

So a longstanding open question (for at least 53 years!) is whether naive policy iteration converges for a broad class of models. Often other attempts to answer this question have only succeeded in proving the convergence of modified versions of the algorithm for restrictive classes of games. The work of \cite{filar1991algorithm} analyzes a modification of the naive policy iteration algorithm. However, the work of \cite{perolat2016softened} shows that their proof hinges on the assumption that the $\scriptL_2$-norm of the optimal Bellman residual is smooth, which is generally not true. A significant recent contribution in \cite{ bertsekas2021distributed} is a modification of naive policy iteration that converges but requires more storage. However, this algorithm does not yet appear to have an extension to the function approximation setting even for Markov games with special structure such as linear Markov games. An extension of \cite{bertsekas2021distributed} can be found in \cite{brahma2022convergence} which studies stochastic and optimistic settings. Our contribution is a modified version of the naive policy iteration algorithm that converges exponentially fast for all discounted-reward/cost zero-sum Markov games. The modification is easy to explain: simply replace the policy improvement step with a lookahead version of policy improvement. Lookahead has been widely used in RL from the early application to backgammon \cite{tesauro1996line} to recent applications such as chess and Go in AlphaZero \cite{silver2017mastering}. But to the best of our knowledge, our result is the first which proves the convergence of naive policy iteration using lookahead. Additionally, we show lookahead has low computational complexity for the class of linear Markov games, which are a natural generalization of linear MDPs, which have been studied extensively recently \cite{agarwal2020flambe, uehara2021representation,zhang2022making}. We note that our result complements the recent results on the benefits of lookahead to improve the convergence properties of MDPs with \cite{annaor, annacdc, winnicki2023convergence} and without function approximation \cite{efroni2018multiple,efroni2018,  efroni2019combine, tomar2020multistep} in other different contexts. For more on lookahead see \cite{efroni2018multiple, efroni2019combine, tomar2020multistep, annaor, annacdc, winnicki2023convergence}. 

In fact, our results are for an algorithm which subsumes policy iteration and value iteration as special cases. Following \cite{perolat2015approximate}, we call the algorithm generalized policy iteration, although the MDP version of the algorithm goes by several names including modified policy iteration \cite{Puterman1978ModifiedPI} and optimistic policy iteration \cite{bertsekastsitsiklis}. 

We also incorporate function approximation into our algorithms. Prior works that extend policy iteration to Markov games have all attempted to extend their work to the setting of function approximation with limited success. The work of \cite{perolat2015approximate} extends the original policy iteration algorithm of \cite{hoffman1966nonterminating}, however, it is limited in two respects. First, the algorithm in \cite{perolat2015approximate} propagates an error bound in the policy evaluation and policy improvement steps. However, their bounds do not explicitly take into account the implementation details of least-squares-based policy evaluation. Hence, analogously to the counter-example in \cite{annaor}, it is unclear whether the bounds in \cite{perolat2015approximate} can be accurately applied in the least squares policy evaluation for games algorithm. Second, the algorithm in \cite{perolat2015approximate} is inefficient for the same reasons that the Hoffman-Karp algorithm is inefficient, i.e., it requires that each policy corresponding to the minimizer be evaluated approximately at each iteration. The work of \cite{lagoudakis2012value} attempts to extend the algorithm of Pollatschek and Avi-Itzhak to the linear value function approximation setting, which, while efficient, does not necessarily converge even in the exact case as shown by the counter-example in \cite{van1978discounted}. Finally, the modification to the algorithm of Pollatschek and Avi-Itzhak, the algorithm in the work of \cite{filar1991algorithm}, has been extended to the function approximation setting in \cite{perolat2016softened}, however, for the same reasons that algorithm also does not converge. We provide algorithms that extend our results to the function approximation setting and give performance bounds which become zero in the special case where there is no function approximation error. In the special case of linear Markov games, the computations do not depend on the size of the state and action spaces and rather depend on the dimension of the feature vectors. 

Finally, to show the applicability of our results, and in particular, to show that it can be combined with learning algorithms for zero-sum Markov games, we present a sample complexity result using the learning phase of the algorithm in the recent work of \cite{zhang2020model} along with our generalized policy iteration algorithm for planning.
Model-based algorithms generally consist of two phases: a learning phase where the probability transition matrix and average reward/cost are estimated, possibly using a generative model, and a planning phase. The results in \cite{zhang2020model} are obtained assuming that there exists an efficient algorithm for planning. The work of \cite{zhang2020model} analyzes the setting where a model is estimated from data and a general algorithm of one's choice including value iteration, policy iteration, etc. is used to find the Nash equilibrium.  Additionally, there are multiple RL algorithms for other versions of model-based multi-agent RL including turn-based Markov games \cite{sidford2020solving}, finite-horizon games \cite{bai2020provable,liu2021sharp}, among others.  
We outline our contributions as follows:

\subsection{Main Contributions:}
\textbf{Convergence Of Generalized Policy Iteration In Markov Games}
The computational difficulties associated with extending the policy iteration algorithm to games has been a longstanding open problem. Several studies shed light on the difficulty of the policy iteration algorithm \cite{hansen2013strategy}. We present a simple modification of the well-studied naïve policy iteration or the algorithm of Pollatschek and Avi-Itzhak \cite{pollatschek1969algorithms} which converges for all discounted, simultaneous-move Markov zero-sum games. Moreover, the algorithm converges exponentially fast.   We remark that our generalized policy iteration algorithm can also be seen as a generalization of both value iteration and policy iteration in games, which is interesting in its own right, even for single player MDPs. 

\textbf{Linear MDPs \& Function Approximation} We then extend the results to incorporate function approximation by studying convergence and scalability to lower dimension linear Markov games, noting recent successful results in making linear MDPs more practical using representation learning techniques \cite{ agarwal2020flambe, uehara2021representation,zhang2022making}. Prior work on approximating lookahead using MCTS shows that exponential computational complexity is inevitable in problems with no structure \cite{shah2020nonasymptotic}. In contrast, our results show that the computational complexity of implementing lookahead only depends on the dimension of the feature vectors if we exploit the linear structure of the problem, which is interesting in its own right, including in the case of single player MDPs. 

We also consider the case where the linear value function representation is not perfect and we provide performance bounds for the algorithm. In the algorithm, it is assumed that returns of the policy evaluation with the lookahead for only several states are generated exactly, and the returns for the rest of states are determined by finding a best fitting parameter. Our bounds are interpretable and in the special case of the tabular setting, i.e., one-hot encoded feature vectors, when all states are evaluated at each iteration, the error of the algorithm is zero. 

\textbf{Learning Value Functions from Noisy Observations} We then consider the case where policy evaluation is performed via observations from a single trajectory, resulting in an unbiased error. We show that a stochastic approximation algorithm for estimating the value function converges to that of the optimal policy up to a function approximation error.

\textbf{Reinforcement Learning} Many papers have studied the RL problem for zero-sum Markov games, in both model-based and model-free settings. The most relevant paper to our setting  \cite{zhang2020model} is agnostic to the planning algorithm used. However, due to the lack of any prior results on the use of policy iteration, the results in that paper will not hold if one were to use naive policy iteration because it is known to not converge in some examples. Here, combining our results with \cite{zhang2020model}, we provide a characterization of the sample complexity involved in model-based learning combined with generalized policy iteration for games.

\section{Related Works}
\textbf{Policy iteration for games} Since the introduction of a mathematical framework for Markov games in \cite{Shapley}, value-based algorithms to obtain a Nash equilibrium for Markov games have been studied extensively, including the early works of \cite{littman1994markov, patekthesis, hu2003nash}. Policy iteration algorithms have been far less successfully studied despite the fact that \cite{van1978discounted} shows that Shapley's value iteration in games is slower in practice than naive policy iteration \cite{pollatschek1969algorithms}. Relevant prior works include \cite{patekthesis, perolat2015approximate} which obtain convergence of policy iteration algorithms for Markov games which require the solution of an MDP at each step, which is computationally burdensome. Furthermore, the work of \cite{pollatschek1969algorithms} following the work of \cite{hoffman1966nonterminating}   proposes an algorithm that is far more computationally efficient but they only show the algorithm converges in specific settings. The works of \cite{filar1991algorithm,breton1986computation} obtain convergence of a variant of the algorithm in \cite{pollatschek1969algorithms} under certain conditions \cite{perolat2016softened}. Recently, the works of \cite{bertsekas2021distributed, brahma2022convergence} study a variant of the algorithm in \cite{pollatschek1969algorithms} that converges, but the dimension of vectors to be stored is quite large and the algorithms have not been shown to be extendible the function approximation setting. 

\textbf{Model-based reinforcement learning in Markov games} The works of \cite{jia2019feature, sidford2020solving} study value-based approaches that assume the use of generative models where any state-action pair can be sampled at any time. Model-based algorithms for Markov games have been widely studied. The work of \cite{zhang2020model} studies a general setting where learning is used to estimate a model and a planning algorithm is applied to obtain the Nash equilibrium policy. Related model-based episodic algorithms that incorporate value iteration for two-player games include \cite{bai2020provable, xie2020learning} in the finite-horizon setting. The work of \cite{liu2021sharp} provides an episodic algorithm where at each iteration there is a planning step which performs an optimistic form of value iteration and there is a learning step where the outcomes of game play in the planning step are used to update the estimate of the model.

\textbf{Policy-based methods for two-player games}
In our work, we study model-based learning, i.e., we learn a model followed by planning. An alternative is to directly learn the policy without learning the model. The work of \cite{daskalakis2020independent} provides a decentralized algorithm for policy gradient methods that converges to a min-max equilibrium when both players independently perform policy gradient. The work of \cite{zhao2022provably} obtains convergence guarantees of natural policy gradient in two-player zero-sum games. 

\textbf{Function approximation methods in multi-agent RL} The work of \cite{lagoudakis2012value} studies linear value function approximation in games where knowledge of the model is assumed. Furthermore, \cite{xie2020learning} studies multi-agent games in linear Markov games, but they only study value iteration in finite-horizon settings. Additionally, the work of \cite{jin2022power} studies episodic learning in multi-agent Markov games with general function approximation. 

\textbf{Planning Oracles For Learning In MDPs} The works of \cite{gheshlaghi2013minimax, agarwal2020model, li2020breaking, jin2020reward} study reinforcement learning in a single agent setting where a planning oracle is used to obtain convergence guarantees. 
\section{Model} \label{section C}
Consider a two-player simultaneous-action zero-sum discounted Markov game. As mentioned in the Introduction, we first consider the planning component of a model-based learning algorithm and later make connection to learning. In the planning setting, the probability transition matrices and reward functions are assumed to be known to both players. The game is characterized by $(\scriptS, \scriptU, \scriptV, P, g, \alpha)$. We denote by $\scriptS$ the finite state space and $|\scriptS|$ the size of the state space. With slight abuse of notation, we say that $\scriptU$ is the finite action space  for the first player (the maximizer) where $|\scriptU|$ denotes the size of the action space for the maximizer. We call $\scriptU(s)$ the set of possible actions at state $s$ where $\scriptU = \cup_{s \in \scriptS} \scriptU(s)$.  Similarly, we call $\scriptV$ the finite action space for the second player (the minimizer) where $|\scriptV|$ is the size of the action space for the minimizer. We denote by $\scriptV(s)$ is the action space at state $s$ where $\scriptV = \cup_{s \in \scriptS} \scriptV(s)$.  We define $P$ as the probability transition kernel where $P(s'|s,u,v)$ is the probability of transitioning from state $s \in \scriptS$ to state $s' \in \scriptS$ when the maximizer takes action $u \in \scriptU (s)$ and the minimizer takes action $v \in \scriptV (s)$. We say that $g: \mathbb{R}^{|\scriptS|\times |\scriptU| \times |\scriptV|} \to [0,1]$ is the reward function (where, for the sake of completeness, we define $g(s, u, v) :=0$ for $u \notin \scriptU(s)$ or $v \notin \scriptV(s)$).

At each time instance $i$, the state of the game is $s_i$ and the maximizer takes action $u_i$ while the minimizer takes action $v_i$. We assume that the players take actions $u_i$ and $v_i$ simultaneously and remark that the setting where the players take moves sequentially, i.e., the setting of turn-based Markov games, is a special case of our setting of simultaneous moves. The maximizer incurs a reward of $g(s_i, u_i, v_i)$ where we assume without loss of generality that $g(s_i,u_i,v_i) \in [0,1]$ while the minimizer incurs a cost of $g(s_i, u_i, v_i)$ (and, hence a reward of $-g(s_i, u_i, v_i)$). By the end of the game, from the perspective of the maximizer, the game receives a discounted sum of the rewards with discount factor $\alpha$ where $0<\alpha<1$, i.e., $\sum_{i=0}^\infty \alpha^i g(s_i, u_i,v_i).$ Meanwhile, from the perspective of the minimizer, the game incurs a discounted sum of the costs, i.e., $\sum_{i=0}^\infty \alpha^i g(s_i, u_i,v_i).$ The objective of maximizer is to take actions $u_i$ to maximize $\sum_{i=0}^\infty \alpha^i g(s_i, u_i,v_i)$ while the objective of the minimizer is to minimize $\sum_{i=0}^\infty \alpha^i g(s_i, u_i,v_i).$

We call a mapping from states to distributions over actions a \textit{policy}, $(\mu,\nu)$. With slight abuse of notation, at each instance $i$, at state $s_i$, the action $u_i$ is selected following a randomized policy $\mu(s_i)$ where $\mu(s_i) \in \Delta(\scriptU(s_i))$ and $\Delta(\scriptU(s_i))$ denotes the set of distributions over actions in $\scriptU(s_i)$. Similarly, at instance $i$, the action $v_i$ is selected following a randomized policy $\nu(s_i)$ where $\nu(s_i) \in \Delta(\scriptV(s_i)).$ 

Given a policy $(\mu,\nu),$ we define the \textit{value function} corresponding to the policy component-wise as follows:$$
    J^{\mu,\nu}(s) = E_{P,\mu,\nu}\Big[\sum_{i=0}^\infty \alpha^i g(s_i, u_i,v_i)|s_0 = s\Big].$$
A pair of policies $(\mu^*, \nu^*)$ is a Nash equilibrium if they satisfy
    $J^{\mu,\nu^*}\leq J^{\mu^*,\nu^*}\leq J^{\mu^*,\nu}
$ for all policies $(\mu,\nu)$. 
It has been shown in \cite{Shapley} that such a Nash equilibrium policy exists for all two-player discounted zero-sum Markov games. We define the value function of the game to be $J^{\mu^*,\nu^*}$ and we will denote it by $J^*$ for convenience.

\section{Preliminaries}\label{prelim}
Consider any policy $(\mu,\nu)$. We will define the probability transition matrix $P_{\mu,\nu} \in \mathbb{R}^{|\scriptS|\times|\scriptS|}$ component-wise where
\begin{align*}
P_{\mu,\nu}(s,s') = \sum_{u \in \scriptU(s)} \sum_{v \in \scriptV(s)} \mu(u)\nu(v) P(s'|s, u,v) \forall (s,s')\in \mathbb{R}^{|\scriptS|\times|\scriptS|}.
\end{align*}
We define the reward function corresponding to policy $(\mu,\nu)$ as $g_{\mu,\nu} \in \mathbb{R}^{|\scriptS|},$ where 
$$g_{\mu,\nu}(s) = \sum_{u \in \scriptU(s)} \sum_{v \in \scriptV(s)} \mu(u)\nu(v) g(s,u,v) \forall s \in \scriptS.$$
Using $g_{\mu,\nu}$ and $P_{\mu,\nu}$, we define the Bellman operator corresponding to policy $(\mu,\nu)$, $T_{\mu,\nu}:\mathbb{R}^{|\scriptS|} \to \mathbb{R}^{|\scriptS|},$ component-wise as follows:
$$T_{\mu,\nu}V(s) := g_{\mu,\nu}(s)+\alpha P_{\mu,\nu} V (s).$$
If operator $T_{\mu,\nu}$ is applied $m$ times to vector $V \in \mathbb{R}^{|\scriptS|},$ then we say that we have performed an $m$-step rollout of the policy $(\mu,\nu)$ and the result $T^m_{\mu,\nu} V$ of the rollout is called the return.
It is well known that $$\norm{T_{\mu,\nu} V - J^{\mu,\nu}}_\infty \leq \alpha \norm{V-J^{\mu,\nu}}_\infty,$$ hence, iteratively applying $T_{\mu,\nu}$ yields convergence to $J^{\mu,\nu}.$ 

We will now give a few well-known properties of the $T_{\mu, \nu}$ operator. First, $T_{\mu, \nu}$ is monotone, i.e., $$V \leq V' \implies T_{\mu, \nu}V \leq T_{\mu, \nu} V'.$$ Herein, we imply that all inequalities hold element-wise. Second, consider the vector $e \in \mathbb{R}^{|\scriptS|}$ where $e(i) = 1 \forall i \in 1, 2, \ldots, |\scriptS|.$ We have that $$T_{\mu, \nu}(V + ce) = T_{\mu, \nu}V + \alpha ce \forall c \in \mathbb{R}.$$

With some algebra, it is easy to see that $T_{\mu,\nu}V$ can also be written component-wise as:
\begin{align}   T_{\mu,\nu}V(s) = \mu(s)^\top A_{V,s} \nu(s) \forall s \in \scriptS, \label{eq: bell mu nu}
\end{align}
where $A_{V,s} \in \mathbb{R}^{|\scriptU(s)|\times|\scriptV(s)|}$ is defined as follows: 
\begin{align}
 A_{V,s}(u,v) := g(s, u, v)+ \alpha \sum_{s' \in \scriptR(s, u, v)} P(s'|s, u,v)V(s'), \label{eq: As}
\end{align} for all $(u,v)  \in (\scriptU(s)\times \scriptV(s))$ where $\scriptR(s, u, v)$ is the set of states for which $P(s'|s,u,v) \neq 0$, i.e., the states that are ``reachable'' from $s$ when taking actions $u$ and $v$. Note that the size of $\scriptR(s,u,v)$ for any $(s, u,v)$ is typically much smaller than the size of the state space. Thus, in order to compute the $m$-step rollout for policy $(\mu,\nu)$ corresponding to vector $V\in \mathbb{R}^{|\scriptS|}$, one can iteratively perform the operations in \eqref{eq: bell mu nu} for all states $s \in \scriptS$ to apply the Bellman operator $T_{\mu,\nu}$ $m$ times.

We define the Bellman optimality operator or Bellman operator $T: \mathbb{R}^{|\scriptS|} \to \mathbb{R}^{|\scriptS|}$ as $$TV = \max_\mu \min_\nu (T_{\mu,\nu} V).$$
We call the resulting $\argmax \argmin$ policy the ``greedy policy,'' i.e.,  $(\mu,\nu) \in \mathcal{G}(V)$ when $$(\mu,\nu) \in \argmax_\mu \argmin_\nu (T_{\mu,\nu} V).$$

Using the notation in \eqref{eq: bell mu nu}, it is easy to see that the Bellman operator at each state $s \in \scriptS$ solves the following matrix game:
\begin{align}TV(s) = \displaystyle\max_{\substack{\mu(s)\in \mathbb{R}^{|\scriptU(s)|} \\  \mu(s)^\top  \textbf{1} =1\\0\leq \mu(s) \leq 1  }} \displaystyle\min_{\substack{\nu(s) \in \mathbb{R}^{|\scriptV(s)|} \\ \nu(s)^\top  \textbf{1} =1\\0\leq \nu(s) \leq 1  }} \mu(s)^\top A_{V,s} \nu(s), \label{eq: bellman op}\end{align}
where $A_{V,s}$ is defined for all $s \in \scriptS$ in \eqref{eq: As} and $\textbf{1}$ the column vector of all 1s. Note that the inequalities $0\leq \mu(s)\leq 1$ and $0\leq \nu(s)\leq 1$ are defined to be component-wise. We define the \textit{greedy policy} $(\mu,\nu)$ corresponding to vector $V$ component-wise where $(\mu(s),\nu(s))$ is the $\argmin \argmax$ in \eqref{eq: bellman op} for all states $ s \in \scriptS$. We remark that the computation of the greedy policy can be obtained by solving a linear program \cite{rubinstein1999experience}. 
Additionally, it is known that $T$ is a pseudo-contraction towards the Nash equilibrium $J^*$ where $$\norm{TJ - J^*}_\infty \leq \alpha \norm{J-J^*}_\infty,$$ hence, iteratively applying $T$ yields convergence to the Nash equilibrium $J^*$ \citep{bertsekastsitsiklis}.

If operator $T$ is applied $H$ times to vector $V \in \mathbb{R}^{|\scriptS|},$ we say that the result, $T^H V$, is the $H$-step ``lookahead'' corresponding to $V$. We call $\scriptL(V)$ the greedy policy corresponding to $T^H V$ the $H$-step lookahead policy, or the lookahead policy, when $H$ is understood. In other words, given an estimate $V$ of the Nash equilibrium, the lookahead policy is the policy $(\mu,\nu)$ such that $$T_{\mu,\nu}(T^{H-1} J)=T(T^{H-1} V).$$ In order to compute the lookahead corresponding to $V \in \mathbb{R}^{|\scriptS|}$ (and the lookahead policy), one can iteratively perform the operations in \eqref{eq: bellman op} for all states $H$ times, obtaining the lookahead policy for each state $s$ by taking the $(\mu(s),\nu(s))$ corresponding to the $\argmax \argmin$  policy in \eqref{eq: bellman op} at the $H$-th iteration of applying $T$. 

Finally, we define the operator $T_{\mu}:\mathbb{R}^{|\scriptS|} \to \mathbb{R}^{|\scriptS|}$ as follows:
\begin{align}
    T_{\mu}V = \min_\nu (T_{\mu,\nu} V). \label{eq:Tmu}
\end{align}
It is known that $T_\mu$ is a maximum norm contraction with discount factor $\alpha$ and with respect to $J^{\mu} \in \mathbb{R}^{|\scriptS|}$ defined as
$J^{\mu} = \min_\nu J^{\mu,\nu},$ i.e.,
\begin{align*}
    \norm{T_{\mu} V - J^{\mu}}_\infty \leq \alpha \norm{V-J^{\mu}}_\infty \forall \text{ } V \in \mathbb{R}^{|\scriptS|}.
\end{align*} Additionally, it is known that $T_{\mu}$ is a monotone operator, i.e., 
\begin{align*}
    J \leq J' \implies T_\mu J \leq T_\mu J'.
\end{align*}

\subsection{Motivation}
The extension of policy iteration to Markov games is given by the following:
\begin{align} 
\nonumber
(\mu_{k+1},\nu_{k+1}) &= \mathcal{G}(V_k) \\
V_{k+1} &= J^{\mu_{k+1}} = T_{\mu_{k+1}}^\infty V_k. \label{eq:PIgames}
\end{align}
A well-known variant of this algorithm involves using an $m$-step return to estimate $J^{\mu_{k+1}}$, i.e., setting $V_{k+1} = T_{\mu_{k+1}}^m V_k.$ 

Notice that in the policy evaluation step, the policy $\mu_{k+1}$ is fixed, $\min_{\nu} J^{\mu_{k+1},\nu}$ is evaluated, and the estimate of the value function is updated to be $\min_{\nu} J^{\mu_{k+1},\nu}$. Note that obtaining $\min_{\nu} J^{\mu_{k+1},\nu}$ requires that an MDP be solved (this is because, since the policy of the maximizer $\mu$ is fixed, only one player, the minimizer, needs to take actions at every iteration to minimize the expected discounted sum of rewards). Hence, while the algorithm converges, in the policy evaluation step of the games setting, unlike the MDP setting, where computing $J^{\nu_{k+1}}$ for the greedy policy $\nu_{k+1}$ can be obtained by inverting a matrix, an MDP must be solved, which makes the algorithm highly inefficient and potentially infeasible. Thus, this  motivates our current work, which is to find an efficient policy iteration algorithm for Markov games. 

\section{Convergence Of Generalized Policy Iteration For Markov Games}
Convergence of a computationally efficient extension of policy iteration for single player systems to two-player games is an open problem \cite{bertsekas2021distributed, patekthesis}. The policy iteration algorithm we consider is given by the following:
\begin{align*}
    \mu_{k+1},\nu_{k+1} &= \scriptL(V_k), \\
    V_{k+1} &= T_{\mu_{k+1},\nu_{k+1}}^m T^{H-1} V_k,
\end{align*} where $\scriptL$ is an $H$-step lookahead policy. Formally, our generalized policy iteration algorithm for two-player games is outlined in Algorithm \ref{alg:alg 2}. 

The algorithm is an iterative process that updates an estimate of the optimal value function at each iteration. At each iteration, there are two steps: the policy improvement step and the policy evaluation step. In the policy improvement step, a new policy to evaluate in the policy evaluation step is determined. The new policy is obtained by computing an $H$-step lookahead policy based on the estimate of the optimal value function.  
In other words, at iteration $k+1,$ the algorithm computes $(\mu_{k+1},\nu_{k+1})$ such that 
\begin{align}
T^H V_k(s) = T_{\mu_{k+1},\nu_{k+1}}T^{H-1}V_k(s) \forall s \in \scriptS
\label{eq:bellman}
\end{align} by solving the linear program in \eqref{eq: bellman op} for all states $s \in \scriptS$ $H$ times. Note that $T^{H-1}V_k$ is computed as a byproduct of determining the lookahead policy, so the estimate of the value function is updated to be $T^{H-1}V_k.$ 

\begin{remark} The use of lookahead policies in the policy improvement step has been used in empirically successful algorithms such as AlphaZero. In practice, lookahead can be implemented efficiently using techniques such as Monte Carlo Tree Search (MCTS).
\hfill $\diamond$
\end{remark}
The policy evaluation step involves  applying the operator $T_{\mu_{k+1},\nu_{k+1}}$ $m$ times to the updated estimate of the optimal value function $\tilde{V}_k = T^{H-1}V_k$ to estimate $J^{\mu_{k+1},\nu_{k+1}}.$ Recall that iteratively applying $T_{\mu_{k+1},\nu_{k+1}}$ to vector $\tilde{V}_k$ yields convergence to $J^{\mu_{k+1},\nu_{k+1}},$ and hence, when $m=\infty$, $V_{k+1} = J^{\mu_{k+1},\nu_{k+1}}.$
Put together, our algorithm can also be written as follows:
\begin{align}
V_{k+1} = T_{\mu_{k+1},\nu_{k+1}}^m T^{H-1}V_k. \label{eq:alt alg}
\end{align}
Note that to apply $T_{\mu_{k+1},\nu_{k+1}}$ to $\tilde{V}_k$ for each state $s \in \scriptS,$ one needs to perform the operations in \eqref{eq: bell mu nu}, and hence, to obtain $T_{\mu_{k+1},\nu_{k+1}}^m\tilde{V}_k,$ i.e., to generate the $m$-return, one needs to perform the computations for all states $s \in \scriptS$ $m$ times. 

We furthermore note that in the \textit{naive policy iteration algorithm} of Pollatschek and Avi-Itzhak, $H$ is set to $1,$ and we simply obtain the ``greedy policy.'' In other words, the naive policy iteration algorithm is given by
\begin{align} 
    \mu_{k+1},\nu_{k+1} \nonumber&= \mathcal{G}(V_k), \\
    V_{k+1} &= T_{\mu_{k+1},\nu_{k+1}}^m V_k,\label{eq:naive}
\end{align} where $\mathcal{G}$ denotes a 1-step greedy policy defined in Section \ref{prelim}. 
We also note that in some ways, our algorithm is a generalization of the naive policy iteration algorithm. 
\begin{algorithm} 
\caption{Generalized PI For Two-Player  Games}\label{alg:alg 2}
\textbf{Input}: $V_0,m, H.$\\ \\
For $k=1,2,\ldots$ \\
 \quad Let $\tilde{V}_k = T^{H-1}V_k$\\
 \quad \quad Let $\mu_{k+1},\nu_{k+1}$ be such that $$\mu_{k+1},\nu_{k+1} \in \argmax_{\mu} \argmin_{\nu} T_{\mu,\nu}\tilde{V}_k.$$\label{step 2 our}\\
 \quad \quad Approximate $J^{\mu_{k+1},\nu_{k+1}}$ as follows: $T_{\mu_{k+1}, \nu_{k+1}}^m \tilde{V}_k.$  \label{step 3 our}
 \\
\quad \quad $V_{k+1} = T_{\mu_{k+1}, \nu_{k+1}}^m \tilde{V}_k.$
\end{algorithm}

\begin{remark} We note that in the case of turn-based Markov games, which are Markov games where the players move sequentially instead of simultaneously, the computations in \eqref{eq: bellman op} that are used to determine the lookahead are far simplified and only involve taking either a maximum or a minimum instead of the $\min \max$ operation in \eqref{eq: bellman op}.
\hfill 
\end{remark}

The proof technique of policy iteration for reward-maximizing single player MDPs  hinges on showing that 
\begin{align}
   J^* \geq \ldots \geq J^{\mu_{k+1}} \geq J^{\mu_k} \geq \ldots \geq J^{\mu_0}.
\label{eq:cascade}
\end{align} 
Recall that policy iteration for single player MDPs is given by the following:
\begin{align*}
    V_{k+1} = J^{\mu_{k+1}}
\end{align*} where $T_{\mu_{k+1}} V_k = \tilde{T} V_k$ and $\tilde{T}$ is defined as follows: 
\begin{align}\tilde{T}V_k = \max_{\mu} T_{\mu} V_k. \label{eq:defT}
\end{align}
To show that $J^{\mu_{k+1}}\geq J^{\mu_k}$ and hence obtain \eqref{eq:cascade}, observe that the following holds:
\begin{align*}
   &T_{\mu_{k+1}} V_k =   \tilde{T} V_k = \tilde{T} J^{\mu_k} \geq J^{\mu_k},
\end{align*} 
where the inequality holds because
\begin{align*}
    \tilde{T} J^{\mu_k} = \max_{\mu} T_{\mu} J^{\mu_k} \geq  T_{\mu_k} J^{\mu_k} = J^{\mu_k}.
\end{align*}
Thus, 
\begin{align}
    T_{\mu_{k+1}} J^{\mu_k} \geq J^{\mu_k}.
\end{align}
Since $T_{\mu_{k+1}}$ is a monotone operator, we can repeatedly apply $T_{\mu_{k+1}}$ to obtain the following:
\begin{align*}
    J^{\mu_{k+1}} \geq \ldots \geq T_{\mu_{k+1}} J^{\mu_k} \geq J^{\mu_k}.
\end{align*}

Notice that the proof of policy iteration for single-player MDPs (as well as its extension to zero-sum Markov games given in \eqref{eq:PIgames}) hinges on the fact that $\tilde{T}J^{\mu_k} \geq J^{\mu_k}$, which naturally follows since $\tilde{T}$ defined in \eqref{eq:defT} involves a maximization over all policies $\mu$. 
In contrast, in the policy improvement step of Algorithm \ref{alg:alg 2} (as well as of naive policy iteration), the greedy policy corresponding to the bellman operator $T$ given in \eqref{eq: bellman op} involves both a maximization as well as a minimization,  i.e., $(\mu_{k+1}, \nu_{k+1}) = \argmax_{\mu} \argmin_{\nu} T_{\mu,\nu} \tilde{V}_k$. Thus, it is not necessarily true that $T_{\mu_{k+1},\nu_{k+1}} \tilde{V}_k \geq \tilde{V}_k.$ In fact, naive policy iteration has been shown to diverge in \cite{van1978discounted, condon1990algorithms}. As such, the question of how to modify the algorithm in \cite{pollatschek1969algorithms} to ensure convergence is an open question where the main challenge is how to overcome the lack of monotonicity in the policy improvement step.
In our algorithm, given in Algorithm \ref{alg:alg 2}, we introduce lookahead policies as opposed to traditionally used greedy policies and use several novel proof ideas to overcome the lack of monotonicity. We remark that other works in the MDP setting also use lookahead policies \cite{efroni2018, efroni2018multiple, efroni2019combine},  however, their arguments often rest on monotonicity arguments which cannot be easily extended in the games setting when the Bellman operator is used.



\subsection{Main Result} We now state our main result, where we prove convergence of Algorithm \ref{alg:alg 2}, a policy iteration algorithm for stochastic games that does not involve solving any MDPs. Our main result hinges on the following assumption on the amount of lookahead in each iteration.

\begin{assumption} \label{assumption 1 games}
$\alpha^{H-1}+2(1+\alpha^m )\frac{\alpha^{H-1}}{1-\alpha}<1.$
\end{assumption} Assumption \ref{assumption 1 games} implies that the lookahead, $H$, must be sufficiently large and also that a large return $m$ can mitigate the amount of lookahead that is needed. We remark that taking steps of lookahead is used in practice such as in algorithms like AlphaZero and that efficient algorithms combined with sampling such as Monte Carlo Tree Search (MCTS) are often employed to perform the lookahead. We also note that the amount of lookahead, $H$, is a parameter of the algorithm, and hence, Assumption \ref{assumption 1 games} is not a restriction on the model, rather an assumption on the parameters of the algorithm. 


\begin{theorem}\label{thm:theorem 1}
The following holds for the iterates of Algorithm \ref{alg:alg 2}:
\begin{align*}
&\norm{V_k-J^*}_\infty  \\&\leq  \Big(\alpha^{H-1}+(1+\alpha^m )\frac{\alpha^{H-1}}{1-\alpha}(1+\alpha)\Big)^k\norm{V_0 - J^*}_\infty.
\end{align*}
Taking limits we can see that
 under Assumption \ref{assumption 1 games},$$ 
V_k \to J^*,
$$ where $V_k$ are the iterates of Algorithm \ref{alg:alg 2} and $J^*$ is the value function.
\hfill $\diamond$
\end{theorem}

The proof of Theorem \ref{thm:theorem 1} can be found in the Appendix.

\begin{remark}(Implications Of Theorem \ref{thm:theorem 1})
We outline the significance of Theorem \ref{thm:theorem 1} as follows:
\begin{itemize}
    \item To the best of our knowledge, our algorithm is the first variant of the well-studied naive policy iteration algorithm that converges without restrictive conditions on the model or additional storage. 
    \item Our algorithm converges exponentially fast to the Nash equilibrium.
    \item Unlike the algorithm in \eqref{eq:PIgames} studied in \cite{patekthesis, perolat2015approximate}, our algorithm does not require that any MDPs be solved at each iteration. 
    \item We remark that the value iteration algorithm for Markov games is a special case of our algorithm, where value iteration for Markov games is given by the following:
    \begin{align*}
        V_{k+1} = T V_k,
    \end{align*} where $T$ is defined in Section \ref{prelim}.
\end{itemize}
\end{remark}

\textbf{Proof Idea} 
Our proof can be found in Appendix \ref{appendix:appendix A}. Our proof techniques do not involve monotonicity and instead hinge on a contraction property towards the Nash equilibrium of the operator $T_{m,H}:\mathbb{R}^{|\scriptS|} \to \mathbb{R}^{|\scriptS|}$  defined as follows:
$$
   T_{m,H}V = T^m_{\mu,\nu}T^{H-1}V,
$$ where $T_{\mu,\nu}(T^{H-1} V)=T(T^{H-1} V).$
Using $T_{m,H}$, our algorithm in \eqref{eq:alt alg} can be written as follows:
\begin{align*}
    V_{k+1} = T_{m,H}(V_k).
\end{align*}
We use the contraction property of this operator in our proofs instead of monotonicity properties used in analyses of policy iteration in single player MDPs \citep{bertsekas2019reinforcement, efroni2019combine}. Doing so allows us to bypass monotonicity complications in games arising from simultaneous minimization and maximization in determining $(\mu_{k+1},\nu_{k+1})$ in \eqref{eq:bellman} as opposed to a single player MDP where actions are taken either to minimize or to maximize, but never to do both at the same time. 
If the lookahead policy $(\mu_{k+1},\nu_{k+1})$ in \eqref{eq:bellman} involved taking only a maximum or a minimum (instead of a joint $\max \min$), one can use monotonicity techniques of \cite{bertsekastsitsiklis, efroni2019combine}, but the main challenge for us is the joint $\max \min$ in the lookahead policy. The key to our proofs lies in the fact that, with sufficient lookahead, the operator $T_{m,H}$ is a contraction. 

In very large systems, function approximation techniques are necessary because of the massive sizes of state spaces. As such, we consider the well-known linear MDP model \cite{Agarwal2019ReinforcementLT} extended to games. We will show that the computations and storage required to determine the Nash equilibrium using Algorithm \eqref{alg:alg 2} in the linear MDP case depends only on the dimension of the feature vectors and not on the size of the state space. 

\section{Linear Value Function Approximation}\label{LinearMDPs}
When the state and actions spaces are very large, function approximation is often necessary. The work of \cite{perolat2015approximate} provides error bounds for the function approximation setting of the algorithm in \eqref{eq:PIgames}. However, even with function approximation, all policies $J^{\mu_{k+1},\nu}$ for all $\nu$ must be evaluated which is inefficient. Convergence difficulties with the naive policy iteration algorithm \cite{van1978discounted} and its variants implies that their extensions to function approximation, such as \cite{perolat2016softened}, suffers from the same difficulty with convergence.  

In the prior section, we assumed that for all states $s \in \scriptS,
V_{k+1}(s)= T_{\mu_{k+1},\nu_{k+1}}^m T^{H-1}V_k(s)$ could be computed. In the case of very large state spaces, function approximation techniques are often employed where $T_{\mu_{k+1},\nu_{k+1}}^m T^{H-1}V_k(s)$ is computed for a fixed subset of states, say, $\scriptD$.
In order to estimate the value function for states not in $\scriptD$, we associate with each state $s \in \scriptS$  a feature vector $\phi(s)\in \mathbb{R}^d$ where typically $d << |\scriptS|$. The matrix comprised of the feature vectors as rows is denoted by $\Phi$. We use those estimates to find the best fitting $\theta \in \mathbb{R}^d$, i.e., 
\begin{align}
    \min_\theta \sum_{s \in D} \Big( (\Phi \theta)(s) - T_{\mu_{k+1},\nu_{k+1}}^m T^{H-1}V_k(s) \Big)^2. \label{eq:find theta}
\end{align} 
The solution to the above minimization problem is denoted by $\theta_{k+1}$. The algorithm then uses $\theta_{k+1}$ to obtain $V_{k+1} = \Phi \theta_{k+1}$. The process then repeats. 
We denote the matrix of feature vectors in $\scriptD$ as rows $\Phi_\scriptD$ and we assume that the rank of $\Phi_\scriptD$ is $d$, i.e., $\Phi_\scriptD$ is full rank. Thus, setting $$\hat{J}^{\mu_{k+1},\nu_{k+1}} := T_{\mu_{k+1},\nu_{k+1}}^m T^{H-1}V_k,$$ we can alternatively rewrite our $\Phi \theta_k$ as follows:
\begin{align}
\Phi \theta_k = \underbrace{\Phi (\Phi_\scriptD^\top \Phi_\scriptD)^{-1}\Phi_\scriptD^\top \scriptP_k}_{=: \scriptM} \hat{J}^{\mu_{k+1},\nu_{k+1}}, \label{eq:use theta}
\end{align} where $\scriptP_k$ is a projection matrix of ones and zeros such that $\scriptP_k \hat{J}^{\mu_{k+1},\nu_{k+1}}$ is a vector whose elements are a subset of the elements in $ \hat{J}^{\mu_{k+1},\nu_{k+1}}$ corresponding to $\scriptD.$ The algorithm is summarized in Algorithm \ref{alg:alg 4 MG}.

\begin{algorithm} 
\caption{Least-Squares Function Approximation Policy Iteration For Markov Games With Lookahead}\label{alg:alg 4 MG}
\textbf{Input}: $\theta_0, m,$ $H,$ feature vectors $\{ \phi(s) \}_{s \in \scriptS}, \phi(s) \in \mathbb{R}^d$  and subset $\scriptD \subseteq \scriptS.$ Here $\scriptD$ is the set of states at which we evaluate the current policy at iteration $k.$
\\\begin{algorithmic}[1] 
\STATE Let $k=0$.
\STATE Let ${\mu_{k+1},\nu_{k+1}},\nu_{k+1}$ be such that $${\mu_{k+1},\nu_{k+1}},\nu_{k+1} \in \argmax_{\mu} \argmin_{\nu} T_{\mu,\nu}T^{H-1}\Phi \theta_k,$$ where the $T$ operator is the Bellman operator.\\
\STATE Compute $\hat{J}^{{\mu_{k+1},\nu_{k+1}}}(s) = T_{{\mu_{k+1},\nu_{k+1}}}^m T^{H-1} (\Phi \theta_k)(s)$ for $i \in \scriptD.$ \\ \label{step 3 alg}
\STATE Choose $\theta_{k+1}$ to solve 
\begin{align}
    \min_\theta \sum_{s \in D} \Big( (\Phi \theta)(s) - \hat{J}^{{\mu_{k+1},\nu_{k+1}}}(s) \Big)^2, \label{step 4 alg}
\end{align} 
where $\Phi$ is a matrix whose rows are the feature vectors.
\STATE Set $k \leftarrow k+1.$ Go to 2.
\end{algorithmic}
\end{algorithm} 
We now present our first main result on the convergence of Algorithm \ref{alg:alg 4 MG}. 

\begin{theorem} \label{theorem 1 MG}
Suppose that $m$ and $H$ satisfy the following:
$$
\underbrace{\alpha^{H-1}+\frac{(\delta_{FV} \alpha^{m+H-1}+\alpha^{H-1})(1+\alpha) }{1-\alpha}}_{=: \kappa} \leq 1, 
$$
where \begin{align*}
\delta_{FV} :=\norm{\scriptM}_\infty
\end{align*}
is a parameter that depends on the feature vectors. 
Then, the following bound holds for iterates $\theta_k$ of Algorithm \ref{alg:alg 4 MG}:
\begin{align*}
&\norm{\Phi \theta_{k}-J^*}_\infty \\&\leq \kappa^k \norm{V_0-J^*}_\infty +\frac{\delta_{app}}{1-\kappa},
\end{align*}
where $\delta_{app}$ is ability of the feature vectors to approximate the policies:
\begin{align*}
    \delta_{app} := \sup_k \norm{J^{\mu_k} - \scriptM  J^{\mu_k}}_\infty.
    \end{align*}
Taking limits as $k\to\infty$:
\begin{align*}
     \limsup_{k\to \infty} \norm{\Phi \theta_k - J^*}_\infty \leq  \frac{\delta_{app}}{1-\kappa}.
\end{align*}
\hfill
$\diamond$
\end{theorem}
The proof of Theorem \ref{theorem 1 MG} can be found in Appendix \ref{appendix: appendix B}. This result was previously published as an invited session paper at IEEE Conference on Decision and Control at the
Marina Bay Sands, Singapore, in 2023.

\begin{remark} (Implications Of Theorem \ref{theorem 1 MG})
Theorem \ref{theorem 1 MG} shows that the performance bounds are small when $\delta_{app}$ is small; recall that  $\delta_{app}$ represents the function approximation error. This suggests that when the value functions can be well approximated by the function approximation, the bound in Theorem \ref{theorem 1 MG} will be small. Additionally, suppose that $\scriptD = \scriptS,$ i.e., when we obtain an estimate of $T_{\mu_{k+1},\nu_{k+1}}^m T^{H-1} V_k(s)$ for all $s \in \scriptS$. Then, $\Phi \theta_k$ converges to $J^*$. This  matches the result of Theorem \ref{thm:theorem 1}. 
The performance bounds also depend on $\delta_{FV}$, which are a function of the choice of feature vectors and states in $\scriptD$. Notice that the coefficient of $\delta_{FV}$ is $\alpha^{m+H-1}$. Thus, sufficiently large $m$ and $H$ can offset the effect of $\delta_{FV}$. For more on $\delta_{FV}$, see the work of \cite{annaor}. This shows that the performance bounds of Theorem \ref{theorem 1 MG} can also be improved with judicious choice of feature vectors and states in $\scriptD.$

\end{remark}
\subsection{Linear MDPs}
We now consider a special case of function approximation, which is the natural extension of linear MDPs to games. We are motivated by the fact that recent works including \cite{agarwal2020flambe,uehara2021representation,zhang2022making} have shown that one can learn linear representations of MDPs efficiently. 

As is standard in the literature on linear MDPs, we assume a model of the following form: 
\begin{align}
g(s,u,v) = \phi(s,u,v)\cdot \theta, \quad P(\cdot|s,u,v) = \phi(s,u,v)\cdot\eta, \label{eq: lin mdp}
\end{align} where each $(s,u,v) \in \scriptS\times \scriptU \times \scriptV$ is associated with a feature vector $\phi(s,u,v) \in \mathbb{R}^d$ where typically $d << |\scriptS|$. We assume $\theta \in \mathbb{R}^d$ and $\eta \in \mathbb{R}^{|\scriptS|\times d}.$ 
These assumptions are described in more detail in \cite{Agarwal2019ReinforcementLT}. We additionally assume that there exists a set of state-actions tuples $\scriptD$ where $\sum_{(s, u, v) \in \scriptD} \phi(s,u,v)\phi(s,u,v)^\top$ is full rank.


We will show that the computations required to converge to $J^*$ in the case of linear MDPs does not depend on the sizes of the state and actions spaces, unlike Algorithms \ref{alg:alg 2} and \ref{alg:alg 4 MG}.

In Algorithm \ref{alg:alg 2}, at iteration $k+1$, the algorithm computes $(\mu_{k+1},\nu_{k+1})$ in \eqref{eq:bellman} by solving the linear program in \eqref{eq: bellman op} for all states $s \in \scriptS$ $H$ times. Then, the algorithm approximates $J^{\mu_{k+1},\nu_{k+1}}$ by applying $T_{\mu_{k+1},\nu_{k+1}}$ $m$ times for each state $s$ where computing $T_{\mu_{k+1},\nu_{k+1}}V(s)$ for any vector $V$ and each state $s$ requires performing the operations in \eqref{eq: bell mu nu}. Hence, the computations required in a single iteration of Algorithm \ref{alg:alg 2} is at least $\mathcal{O}(Hm|\scriptS|^2|\scriptU|^2|\scriptV|^2),$ which is infeasible when the state and action spaces are very large. 

We now show how to obtain $V_{k+1}$ in a way that does not depend on the sizes of state and action spaces. In the case of linear MDPs, any vector $V \in \mathbb{R}^{|\scriptS|}$ can be parameterized by some $\beta \in \mathbb{R}^d$ as follows. Consider any state-actions tuple $(s, u, v)  \in \scriptS\times \scriptU \times \scriptV$. Then, it can be easily shown that $A_{V,s}$ defined in \eqref{eq: As} can be written in the following form:
\begin{align}
    A_{V,s}(u,v) = \phi(s, u, v)^\top \beta \label{eq: AVs}
\end{align} for some $\beta$ \cite{Agarwal2019ReinforcementLT}.

Under this parameterization, we will show how to obtain $\beta'$ corresponding to $T_{\mu,\nu} V$ from $\beta$ corresponding to $V$ for any $V \in \mathbb{R}^{|\scriptS|}$ in a way that the number of computations does not depend on the size of the state space. We will then extend the result to obtain $\beta'$ corresponding to $TV$. First, consider a set of state-actions tuples $\scriptD$ where $\sum_{(s, u, v) \in \scriptD} \phi(s,u,v)\phi(s,u,v)^\top$ is full rank. For $(s,u,v) \in \scriptD$, we can directly compute $A_{T_{\mu,\nu}V,s}(u,v)$, i.e.,
\begin{align}
    \nonumber A_{T_{\mu,\nu}V,s}(u,v) \nonumber \nonumber&= g(s, u, v)+ \alpha \sum_{s' \in \scriptR(s, u, v)} P(s'|s, u,v){T_{\mu,\nu}V,s}(s')\\
    &= g(s, u, v)+ \alpha \sum_{s' \in \scriptR(s, u, v)} P(s'|s, u,v) (\mu(s')^\top A_{V,s'}\nu(s')),\label{eq:A fn ap}
\end{align} where $\scriptR(s,u,v)$ denotes the set of states reachable from $(s,u,v)$ and $A_{V,s'}$ can be constructed using \eqref{eq: AVs}. 

Then, using the resulting $A_{T_{\mu,\nu}V,s}(u,v)$ for $(s,u,v) \in \scriptD$, we can easily obtain an appropriate $\beta'$ by performing a least squares minimization. More precisely, since $$A_{T_{\mu,\nu}V,s}(u,v) = \phi(s,u,v)^\top \beta' \quad \forall (s,u,v) \in \scriptS\times\scriptU(s)\times \scriptV(s),$$ it is sufficient to compute $A_{T_{\mu,\nu}V,s}(u,v)$ for $(s, u, v) \in \scriptD$ and perform a least squares minimization to determine $\beta'$, i.e., 
\begin{align}
    \beta' := \argmin_{(s,u,v)} \sum_{(s, u, v)\in \scriptD}(A_{T_{\mu,\nu}V,s}(u,v) - \phi(s,u,v)^\top \beta')^2.
\end{align} Recall that a unique minimizer exists because of the full rank condition in the definition of $\scriptD.$

The above minimization produces the weight vector $\beta'$ associated with $T_{\mu,\nu}V.$ In a similar manner, one can also obtain the weight vector corresponding to $TV$. The only modification is that instead of computing $\mu(s)^\top A_{TV,s} \nu(s)$, one needs to instead obtain $\min_{\mu} \max_{\nu} \mu(s)^\top A_{T_{\mu,\nu} V,s} \nu(s)$.

Put together, the above shows that in order to obtain a sequence of $\beta_k$ that parameterize the sequence $V_k$ in Algorithm \ref{alg:alg 2}, it is only necessary to perform the sequence of computations described above which do not directly depend on the size of the state space, but only depends on the number of states that can be reached from a given state which is likely to be considerably smaller. 

We summarize the above discussion with the following Proposition.
\begin{proposition}
In general, $H |\scriptS|$ matrix games are required to be solved at each iteration of Algorithm \ref{alg:alg 2}. However, in the special case of linear MDPs, the computations of Algorithm \ref{alg:alg 2} require only that $H \sum_{(s,u,v)\in \scriptD} |\scriptR(s,u,v)|$ matrix games be solved at each iteration when $\scriptD$ is full rank. Additionally, the rest of the computations required to obtain the Nash equilibrium do not depend on the sizes of the state and action spaces.
\hfill $\diamond$
\end{proposition}
\begin{remark} We note that the computations are further simplified in the case of turn-based Markov games, where players take turns to execute actions in alternating time steps. This is because in the setting of turn-based MDPs, only one player at a time is performing either a maximization or a minimization.
\hfill \end{remark}
\section{Learning Value Functions}
Note that in the previous section, we assume that exact estimates of $\hat{J}^{{\mu_{k+1},\nu_{k+1}}}(s) = T_{{\mu_{k+1},\nu_{k+1}}}^m T^{H-1} (\Phi \theta_k)(s)$ for $s \in \scriptD$ are available at each iteration. However, this is not possible in general. On the other hand, by observing a single trajectory of (state, action, reward) triplets under the policy $(\mu_{k+1},\nu_{k+1}),$ it would be possible to obtain an unbiased estimate of  $T_{{\mu_{k+1},\nu_{k+1}}}^m T^{H-1} (\Phi \theta_k)(s)+w_{k+1}(s)$; see \cite{winnicki2023convergence} for details in a single-player setting. Such an unbiased estimate can be denoted by $T_{{\mu_{k+1},\nu_{k+1}}}^m T^{H-1} (\Phi \theta_k)(s)+w_{k+1}(s)$, where $w_{k+1}(s)$ is a conditionally unbiased noise term. Since the value function estimate is noisy, we incorporate stochastic approximation techniques to combine noisy policy evaluation with the general policy iteration algorithm described in the previous sections. Our proposed algorithm is described in Algorithm \ref{alg:SA MG}.
\begin{algorithm}
\caption{Least Squares Function Approximation For Policy Iteration In Markov Games With Unbiased Noise and Lookahead}
\label{alg:SA MG}
\textbf{Input}: $\theta_0,m, H$ feature vectors $\{ \phi(s) \}_{i \in \scriptS}, \phi(s) \in \mathbb{R}^d$  and subsets $\scriptD_k  \subseteq \scriptS, k = 0, 1, \ldots.$ Here $\scriptD_k$ is the set of states visited by a trajectory corresponding to the current policy at iteration $k.$\\
\begin{algorithmic}[1] 
\STATE Let $k=0$.
\STATE Let ${\mu_{k+1},\nu_{k+1}}$ be such that $${\mu_{k+1},\nu_{k+1}},\nu_{k+1} \in \argmax_{\mu} \argmin_{\nu} T_{\mu,\nu}T^{H-1}\Phi \theta_k,$$ where the $T$ operator is the Bellman operator.\\
\STATE    Compute $$\hat{J}^{{\mu_{k+1},\nu_{k+1}}}(s) =  T_{{\mu_{k+1},\nu_{k+1}}}^m T^{H-1}\Phi \theta_k(s)+w_{k}(s)$$ for $s \in \scriptD_k$ and set $\hat{J}^{{\mu_{k+1},\nu_{k+1}}}(s)=0$ for $s \notin \scriptD_k.$ \\ 
\STATE Choose $\theta_{k+1}$ to solve 
\begin{align}
    &\min_\theta \norm{(\scriptP_{1, k}\Phi) \theta - \scriptP_{2, k}\hat{J}^{{\mu_{k+1},\nu_{k+1}}}}_2^2, 
\end{align} \\
where $\Phi$ is a matrix whose rows are the feature vectors ($\scriptP_{1, k}$ and $\scriptP_{2, k}$ are projection matrices defined in the main body of the paper).
\STATE \begin{align}
    \theta_{k+1} = (1-\gamma_k)\theta_k + \gamma_k (\theta_{k+1}). 
\end{align}
\STATE Set $k \leftarrow k+1.$ Go to 2.
\end{algorithmic}
\end{algorithm}

Defining $$V_k := \Phi \theta_k,$$ the iterates in Algorithm \ref{alg:SA MG} can be written as follows:
\begin{align}
    V_{k+1}&= (1-\gamma_k)V_k + \gamma_k (\Phi \theta_{k+1})\nonumber\\&= (1-\gamma_k)V_k + \gamma_k (\Phi(\scriptP_{1,k}\Phi)^+ \hat{J}^{{\mu_{k+1},\nu_{k+1}}})\nonumber\\&\nonumber= (1-\gamma_k)V_k \\&+ \gamma_k (\underbrace{\Phi(\scriptP_{1,k}\Phi)^+ \scriptP_{2,k}}_{=: \scriptM_k} (T_{{\mu_{k+1},\nu_{k+1}}}^m T^{H-1}V_k+w_k)),\label{eq: iter alg 3 unbiased full}
\end{align}
where $(\scriptP_{1,k}\Phi)^+$ is the Moore-Penrose inverse of $\scriptP_{1,k}\Phi$ and $\scriptP_{1,k}$ is a matrix of zeros and ones such that rows of $\scriptP_{1,k}\Phi$ correspond to feature vectors associated with states in $\scriptD_k$ and $\scriptP_{2,k}(\hat{J}^{{\mu_{k+1},\nu_{k+1}}})$ is a vector whose elements are a subset of the elements of $\hat{J}^{{\mu_{k+1},\nu_{k+1}}}$ corresponding to $\scriptD_k$.
We define the term $\delta_{FV}'$ associated with our feature vectors $\phi(s)\forall s \in \scriptS$ as follows:
\begin{align}\delta'_{FV}:=\sup_k \norm{\scriptM_k}_\infty.
\label{eq:def delta MG}
\end{align}
Using $\delta'_{FV}$, we now give Assumption \ref{assumption 2 MG}.
\begin{assumption}\label{assumption 2 MG}
\begin{enumerate}[label=(\alph*)]

\item The starting state of the trajectory at each instance is drawn from a fixed distribution, $p$, where $p(i)>0\forall i \in \scriptS.$
\item  $$\delta'_{FV} \alpha^{m+H-1} \frac{1+\alpha}{1-\alpha}+\frac{2\alpha^{H-1}}{1-\alpha} < 1.$$ 
\item $\sum_{i=0}^\infty \gamma_i = \infty$. Also, $\sum_{i=0}^\infty \gamma_i^2 < \infty.$
\end{enumerate}
\hfill 
$\diamond$
\end{assumption}
We make several remarks on our assumptions:
\begin{enumerate}[label=(\alph*)]
    \item The first assusmption is what is denoted as ``exploring starts'' (see \cite{sutton2018reinforcement}), and guarantees for all states to be selected infinitely many times. We note that it is straightforward to extend our results to any initial distribution as long as the probability of visiting any state is lower bounded by a constant. In particular, we do not require a fixed probability distribution for the initial state. 
    \item We assume the lookahead is sufficiently large, see previous sections for more on lookahead.
    \item The stepsizes are square summable and sum to infinity, which allows for noise averaging.
\end{enumerate}

We now provide our main performance bounds for Algorithm \ref{alg:SA MG}: 
\begin{theorem} \label{thm:theorem 3 unbiased full} 
Under Assumption \ref{assumption 2 MG},
the iterates obtained in \eqref{eq: iter alg 3 unbiased full} almost surely have the following property:
\begin{align*}
&\limsup_{k \to \infty} \norm{V_k-J^*}_\infty \leq \frac{\delta'_{app}}{1-\Big( \alpha^{H-1}+(1+\alpha)\frac{(1+ \alpha^m \delta'_{FV})\alpha^{H-1}}{1-\alpha} \Big)},
\end{align*}
where $\delta'_{app}$ is ability of the feature vectors to approximate the policies:
\begin{align}
 \nonumber&\delta'_{app} := \\&\sup_k E[\norm{\scriptM_k( J^{{\mu_{k+1},\nu_{k+1}}}+ w_k)-(J^{{\mu_{k+1},\nu_{k+1}}}+ w_k)}_\infty|\scriptF_k]
\end{align} and $\delta_{FV}'$ is a function of the feature vectors
\begin{align*}\delta'_{FV}:=\sup_k \norm{\scriptM_k}_\infty.
\end{align*}\hfill$\diamond$
\end{theorem}
The proof of Theorem \ref{thm:theorem 3 unbiased full} can be found in Appendix \ref{appendix: appendix C}. This result was previously published as an invited session paper at IEEE Conference on Decision and Control at the
Marina Bay Sands, Singapore, in 2023.

\begin{remark} \textbf{Interpretation of Theorem \ref{thm:theorem 3 unbiased full}} 
Analogously to Theorem \ref{theorem 1 MG}, Theorem \ref{thm:theorem 3 unbiased full} shows that the performance bound is mostly based on the ability of the feature vectors to represent unbiased estimates of the value functions. Without feature vectors (i.e., when feature vectors are simply unit vectors) and trajectories from all states are obtained (i.e., in the special case where the Markov chains induced by all policies are irreducible and infinitely long trajectories are obtained), the error becomes zero. 
\end{remark}

\section{Application To RL} \label{section G}
In multi-agent model-based reinforcement learning algorithms, the learning component has been extensively studied, and hence many straightforward extensions of our work exist to involve learning in model-based settings. We will provide one such example of an algorithm based on the learning algorithm for model-based multi-agent reinforcement learning in the work of \cite{zhang2020model}. The algorithm we study can be described as follows.
The algorithm assumes knowledge of the cost/reward function (this setting is called the \textit{cost/reward-aware} setting) \cite{zhang2020model} as well as access to a generator, which, at any iteration, for any state-actions tuple $(s,u,v)$ can sample from the distribution $P(\cdot|s, u,v)$ and obtain the next state. For each state-actions tuple $(s,u,v),$ the algorithm obtains $N$ samples and, based on the samples constructs an estimate of the probability transition matrix in the following manner: $\hat{P}(s'|s,u,v) := \frac{\text{count}(s',s,u,v)}{N}.$ Using $\hat{P}$ and the known cost/reward function, the algorithm finds the Nash equilibrium policy using Algorithm \ref{alg:alg 2}, $(\hat{\mu},\hat{\nu})$. 
The following theorem gives a bound on the sample and computational complexity required to achieve an error bound on $\norm{Q^{\hat{\mu},\hat{\nu}}-Q^*}_\infty$ in linear turn-based Markov games where $\phi(s, u,v) \in \mathbb{R}^d$ and the number of reachable states from state-actions tuples in $\scriptD$ is $r.$

\begin{theorem}\label{thm:theorem 3}
    Consider a linear turn-based Markov game and any $\epsilon, \delta, \epsilon_{opt}>0$ with $\epsilon \in (0, 1/(1-\alpha)^{1/2}]$. When the number of samples of each state-actions tuple is at least $N$ and the number of computations that are made in the planning step where the Nash equilibrium policy is determined based on the model inferred from the samples is at least $C$ where
\begin{align*}
 &N \geq \frac{c \alpha \log\big[ c |\scriptS||\scriptU||\scriptV|(1-\alpha)^{-2}\delta^{-1}\big]}{(1-\alpha)^3 \epsilon^2}, \\&C \geq \frac{c  m H \log\Big[ \frac{1}{\epsilon_{opt}(1-\alpha)}\Big]}{\log[\frac{1}{\tilde{\alpha}}]}\Bigg[ d[2r+1]+d^3/3+r|\scriptA|_{max}^2 d \Bigg]
\end{align*} where $c$ is a constant, $\tilde{\alpha}=\alpha^{H-1}+(1+\alpha^m)\frac{\alpha^{H-1}}{1-\alpha}(1+\alpha),$ $|\scriptU|$ and $|\scriptV|$ are the numbers of the total actions available to players 1 and 2, and $|\scriptA|_{max}$ is the largest number of actions available at a state, it holds that with probability at least $1-\delta,$ 
$\norm{Q^{{\hat{\mu},\hat{\nu}}}-Q^*}_\infty \leq \frac{2\epsilon}{3}+\frac{5 \alpha \epsilon_{opt}}{1-\alpha}, \norm{\hat{Q}^{\hat{\mu},\hat{\nu}}-Q^*}_\infty \leq \epsilon + \frac{9\alpha \epsilon_{opt}}{1-\alpha}.$

\hfill $\diamond$
\end{theorem}

The proof of Theorem \ref{thm:theorem 3} uses the results of \cite{zhang2020model} and extensions from Theorem \ref{thm:theorem 1} and can be found in the Appendix. Theorem \ref{thm:theorem 3} overall gives a bound on the the error of the learning algorithm as a function of the number of computations in the planning step and the number of samples in the learning step. We note that convergence of the learning algorithm in the present section does not require solving an MDP at each iteration the way that many model-based policy iteration algorithms do. In some ways, Theorem \ref{thm:theorem 3} provides a trade-off between sample complexity in the learning step and computational complexity in the planning step. 

\section{Conclusions}
In our work, we study the model-based learning problem and focus on the planning step of the problem using known learning results to provide results for model-based policy iteration for two-player zero-sum simultaneous discounted games. The main result of the paper shows that the naive policy iteration algorithm of Avi-Itzhak and Pollatshek converges if lookahead is used in the policy improvement step for both players. This adds to the body of recent literature which shows that lookahead is an important component of the bag of tricks needed to ensure that RL algorithms converge \cite{winnicki2023convergence, annaor}. The fact that lookahead provides an $\mathcal{O}(\alpha^H)$ approximate solution to the optimal policy is a somewhat trivial statement; what we have show here and in \cite{annaor, winnicki2023convergence} is a much stronger statement: lookahead leads to convergence of algorithms while there are counterexamples to show that the corresponding algorithms without lookahead may not converge. Further, we have shown that, in the case of linear MDPs, lookahead is not computationally expensive to implement.

One interesting direction for future work includes extending the results to the stochastic shortest path games problem \cite{patekthesis}.


\bibliographystyle{informs2014} 
\bibliography{refs2.bib}

\begin{thebibliography}{59}
\providecommand{\natexlab}[1]{#1}
\providecommand{\url}[1]{\texttt{#1}}
\providecommand{\urlprefix}{URL }

\bibitem[{Agarwal et~al.(2019)Agarwal, Jiang, Kakade, \protect\BIBand{}
  Sun}]{Agarwal2019ReinforcementLT}
Agarwal A, Jiang N, Kakade SM, Sun W (2019) Reinforcement learning: Theory and
  algorithms. \emph{CS Dept., UW Seattle, Seattle, WA, USA, Tech. Rep} 10--4.

\bibitem[{Agarwal et~al.(2020{\natexlab{a}})Agarwal, Kakade, Krishnamurthy,
  \protect\BIBand{} Sun}]{agarwal2020flambe}
Agarwal A, Kakade S, Krishnamurthy A, Sun W (2020{\natexlab{a}}) Flambe:
  Structural complexity and representation learning of low rank mdps.
  \emph{Advances in neural information processing systems} 33:20095--20107.

\bibitem[{Agarwal et~al.(2020{\natexlab{b}})Agarwal, Kakade, \protect\BIBand{}
  Yang}]{agarwal2020model}
Agarwal A, Kakade S, Yang LF (2020{\natexlab{b}}) Model-based reinforcement
  learning with a generative model is minimax optimal. \emph{Conference on
  Learning Theory}, 67--83 (PMLR).

\bibitem[{Bai \protect\BIBand{} Jin(2020)}]{bai2020provable}
Bai Y, Jin C (2020) Provable self-play algorithms for competitive reinforcement
  learning. \emph{International conference on machine learning}, 551--560
  (PMLR).

\bibitem[{Bertsekas(2021)}]{bertsekas2021distributed}
Bertsekas D (2021) Distributed asynchronous policy iteration for sequential
  zero-sum games and minimax control. \emph{arXiv preprint arXiv:2107.10406} .

\bibitem[{Bertsekas \protect\BIBand{} Tsitsiklis(1996)}]{bertsekastsitsiklis}
Bertsekas D, Tsitsiklis J (1996) \emph{Neuro-dynamic Programming} (Athena
  Scientific), ISBN 9781886529106.

\bibitem[{Bertsekas(2019)}]{bertsekas2019reinforcement}
Bertsekas DP (2019) \emph{Reinforcement learning and optimal control} (Athena
  Scientific Belmont, MA).

\bibitem[{Brahma et~al.(2022)Brahma, Bai, Do, \protect\BIBand{}
  Doan}]{brahma2022convergence}
Brahma S, Bai Y, Do DA, Doan TT (2022) Convergence rates of asynchronous policy
  iteration for zero-sum markov games under stochastic and optimistic settings.
  \emph{2022 IEEE 61st Conference on Decision and Control (CDC)}, 3493--3498
  (IEEE).

\bibitem[{Breton et~al.(1986)Breton, Filar, Haurle, \protect\BIBand{}
  Schultz}]{breton1986computation}
Breton M, Filar JA, Haurle A, Schultz TA (1986) \emph{On the computation of
  equilibria in discounted stochastic dynamic games} (Springer).

\bibitem[{Condon(1990)}]{condon1990algorithms}
Condon A (1990) On algorithms for simple stochastic games. \emph{Advances in
  computational complexity theory} 13:51--72.

\bibitem[{Daskalakis et~al.(2020)Daskalakis, Foster, \protect\BIBand{}
  Golowich}]{daskalakis2020independent}
Daskalakis C, Foster DJ, Golowich N (2020) Independent policy gradient methods
  for competitive reinforcement learning. \emph{Advances in neural information
  processing systems} 33:5527--5540.

\bibitem[{Efroni et~al.(2018{\natexlab{a}})Efroni, Dalal, Scherrer,
  \protect\BIBand{} Mannor}]{efroni2018}
Efroni Y, Dalal G, Scherrer B, Mannor S (2018{\natexlab{a}}) Beyond the one
  step greedy approach in reinforcement learning. \emph{CoRR} abs/1802.03654,
  \urlprefix\url{http://arxiv.org/abs/1802.03654}.

\bibitem[{Efroni et~al.(2018{\natexlab{b}})Efroni, Dalal, Scherrer,
  \protect\BIBand{} Mannor}]{efroni2018multiple}
Efroni Y, Dalal G, Scherrer B, Mannor S (2018{\natexlab{b}}) Multiple-step
  greedy policies in online and approximate reinforcement learning. \emph{arXiv
  preprint arXiv:1805.07956} .

\bibitem[{Efroni et~al.(2019)Efroni, Dalal, Scherrer, \protect\BIBand{}
  Mannor}]{efroni2019combine}
Efroni Y, Dalal G, Scherrer B, Mannor S (2019) How to combine tree-search
  methods in reinforcement learning. \emph{Proceedings of the AAAI Conference
  on Artificial Intelligence}, volume~33, 3494--3501.

\bibitem[{Filar \protect\BIBand{} Tolwinski(1991)}]{filar1991algorithm}
Filar JA, Tolwinski B (1991) \emph{On the Algorithm of Pollatschek and
  Avi-ltzhak} (Springer).

\bibitem[{Gheshlaghi~Azar et~al.(2013)Gheshlaghi~Azar, Munos, \protect\BIBand{}
  Kappen}]{gheshlaghi2013minimax}
Gheshlaghi~Azar M, Munos R, Kappen HJ (2013) Minimax pac bounds on the sample
  complexity of reinforcement learning with a generative model. \emph{Machine
  learning} 91:325--349.

\bibitem[{Gu et~al.(2017)Gu, Holly, Lillicrap, \protect\BIBand{}
  Levine}]{gu2017deep}
Gu S, Holly E, Lillicrap T, Levine S (2017) Deep reinforcement learning for
  robotic manipulation with asynchronous off-policy updates. \emph{2017 IEEE
  international conference on robotics and automation (ICRA)}, 3389--3396
  (IEEE).

\bibitem[{Hansen et~al.(2013)Hansen, Miltersen, \protect\BIBand{}
  Zwick}]{hansen2013strategy}
Hansen TD, Miltersen PB, Zwick U (2013) Strategy iteration is strongly
  polynomial for 2-player turn-based stochastic games with a constant discount
  factor. \emph{Journal of the ACM (JACM)} 60(1):1--16.

\bibitem[{Hoffman \protect\BIBand{} Karp(1966)}]{hoffman1966nonterminating}
Hoffman AJ, Karp RM (1966) On nonterminating stochastic games. \emph{Management
  Science} 12(5):359--370.

\bibitem[{Hu \protect\BIBand{} Wellman(2003)}]{hu2003nash}
Hu J, Wellman MP (2003) Nash q-learning for general-sum stochastic games.
  \emph{Journal of machine learning research} 4(Nov):1039--1069.

\bibitem[{Jia et~al.(2019)Jia, Yang, \protect\BIBand{} Wang}]{jia2019feature}
Jia Z, Yang LF, Wang M (2019) Feature-based q-learning for two-player
  stochastic games. \emph{arXiv preprint arXiv:1906.00423} .

\bibitem[{Jin et~al.(2020)Jin, Krishnamurthy, Simchowitz, \protect\BIBand{}
  Yu}]{jin2020reward}
Jin C, Krishnamurthy A, Simchowitz M, Yu T (2020) Reward-free exploration for
  reinforcement learning. \emph{International Conference on Machine Learning},
  4870--4879 (PMLR).

\bibitem[{Jin et~al.(2022)Jin, Liu, \protect\BIBand{} Yu}]{jin2022power}
Jin C, Liu Q, Yu T (2022) The power of exploiter: Provable multi-agent rl in
  large state spaces. \emph{International Conference on Machine Learning},
  10251--10279 (PMLR).

\bibitem[{Lagoudakis \protect\BIBand{} Parr(2012)}]{lagoudakis2012value}
Lagoudakis M, Parr R (2012) Value function approximation in zero-sum markov
  games. \emph{arXiv preprint arXiv:1301.0580} .

\bibitem[{Li et~al.(2020)Li, Wei, Chi, Gu, \protect\BIBand{}
  Chen}]{li2020breaking}
Li G, Wei Y, Chi Y, Gu Y, Chen Y (2020) Breaking the sample size barrier in
  model-based reinforcement learning with a generative model. \emph{Advances in
  neural information processing systems} 33:12861--12872.

\bibitem[{Littman(1994)}]{littman1994markov}
Littman ML (1994) Markov games as a framework for multi-agent reinforcement
  learning. \emph{Machine learning proceedings 1994}, 157--163 (Elsevier).

\bibitem[{Liu et~al.(2021)Liu, Yu, Bai, \protect\BIBand{} Jin}]{liu2021sharp}
Liu Q, Yu T, Bai Y, Jin C (2021) A sharp analysis of model-based reinforcement
  learning with self-play. \emph{International Conference on Machine Learning},
  7001--7010 (PMLR).

\bibitem[{Mnih et~al.(2016)Mnih, Badia, Mirza, Graves, Lillicrap, Harley,
  Silver, \protect\BIBand{} Kavukcuoglu}]{DBLP:journals/corr/MnihBMGLHSK16}
Mnih V, Badia AP, Mirza M, Graves A, Lillicrap TP, Harley T, Silver D,
  Kavukcuoglu K (2016) Asynchronous methods for deep reinforcement learning.
  \emph{CoRR} abs/1602.01783, \urlprefix\url{http://arxiv.org/abs/1602.01783}.

\bibitem[{Ozdaglar et~al.(2021)Ozdaglar, Sayin, \protect\BIBand{}
  Zhang}]{ozdaglar2021independent}
Ozdaglar A, Sayin MO, Zhang K (2021) Independent learning in stochastic games.
  \emph{arXiv preprint arXiv:2111.11743} .

\bibitem[{Patek(1997)}]{patekthesis}
Patek SD (1997) \emph{Stochastic and shortest path games: theory and
  algorithms}. Ph.D. thesis, Massachusetts Institute of Technology.

\bibitem[{P{\'e}rolat et~al.(2016)P{\'e}rolat, Piot, Geist, Scherrer,
  \protect\BIBand{} Pietquin}]{perolat2016softened}
P{\'e}rolat J, Piot B, Geist M, Scherrer B, Pietquin O (2016) Softened
  approximate policy iteration for markov games. \emph{International Conference
  on Machine Learning}, 1860--1868 (PMLR).

\bibitem[{Perolat et~al.(2015)Perolat, Scherrer, Piot, \protect\BIBand{}
  Pietquin}]{perolat2015approximate}
Perolat J, Scherrer B, Piot B, Pietquin O (2015) Approximate dynamic
  programming for two-player zero-sum markov games. \emph{International
  Conference on Machine Learning}, 1321--1329 (PMLR).

\bibitem[{Pollatschek \protect\BIBand{}
  Avi-Itzhak(1969)}]{pollatschek1969algorithms}
Pollatschek M, Avi-Itzhak B (1969) Algorithms for stochastic games with
  geometrical interpretation. \emph{Management Science} 15(7):399--415.

\bibitem[{Puterman \protect\BIBand{} Shin(1978)}]{Puterman1978ModifiedPI}
Puterman M, Shin MC (1978) Modified policy iteration algorithms for discounted
  markov decision problems. \emph{Management Science} 24:1127--1137.

\bibitem[{Qu et~al.(2022)Qu, Wierman, \protect\BIBand{} Li}]{qu2022scalable}
Qu G, Wierman A, Li N (2022) Scalable reinforcement learning for multiagent
  networked systems. \emph{Operations Research} 70(6):3601--3628.

\bibitem[{Rubinstein(1999)}]{rubinstein1999experience}
Rubinstein A (1999) Experience from a course in game theory: pre-and postclass
  problem sets as a didactic device. \emph{Games and Economic Behavior}
  28(1):155--170.

\bibitem[{Shah et~al.(2020)Shah, Xie, \protect\BIBand{}
  Xu}]{shah2020nonasymptotic}
Shah D, Xie Q, Xu Z (2020) Non-asymptotic analysis of monte carlo tree search.
  \emph{Abstracts of the 2020 SIGMETRICS/Performance Joint International
  Conference on Measurement and Modeling of Computer Systems}, 31--32.

\bibitem[{Shalev-Shwartz et~al.(2016)Shalev-Shwartz, Shammah, \protect\BIBand{}
  Shashua}]{shalev2016safe}
Shalev-Shwartz S, Shammah S, Shashua A (2016) Safe, multi-agent, reinforcement
  learning for autonomous driving. \emph{arXiv preprint arXiv:1610.03295} .

\bibitem[{Shapley(1953)}]{Shapley}
Shapley LS (1953) Stochastic games. \emph{Proceedings of the National Academy
  of Sciences} 39(10):1095--1100, ISSN 0027-8424,
  \urlprefix\url{http://dx.doi.org/10.1073/pnas.39.10.1095}.

\bibitem[{Sidford et~al.(2020)Sidford, Wang, Yang, \protect\BIBand{}
  Ye}]{sidford2020solving}
Sidford A, Wang M, Yang L, Ye Y (2020) Solving discounted stochastic two-player
  games with near-optimal time and sample complexity. \emph{International
  Conference on Artificial Intelligence and Statistics}, 2992--3002 (PMLR).

\bibitem[{Silver et~al.(2016)Silver, Huang, Maddison, Guez, Sifre, Van
  Den~Driessche, Schrittwieser, Antonoglou, Panneershelvam, Lanctot
  et~al.}]{silver2016mastering}
Silver D, Huang A, Maddison CJ, Guez A, Sifre L, Van Den~Driessche G,
  Schrittwieser J, Antonoglou I, Panneershelvam V, Lanctot M, et~al. (2016)
  Mastering the game of go with deep neural networks and tree search.
  \emph{Nature} 529(7587):484--489.

\bibitem[{Silver et~al.(2017{\natexlab{a}})Silver, Hubert, Schrittwieser,
  Antonoglou, Lai, Guez, Lanctot, Sifre, Kumaran, Graepel, Lillicrap, Simonyan,
  \protect\BIBand{} Hassabis}]{silver2017shoji}
Silver D, Hubert T, Schrittwieser J, Antonoglou I, Lai M, Guez A, Lanctot M,
  Sifre L, Kumaran D, Graepel T, Lillicrap TP, Simonyan K, Hassabis D
  (2017{\natexlab{a}}) Mastering chess and shogi by self-play with a general
  reinforcement learning algorithm. \emph{CoRR} abs/1712.01815,
  \urlprefix\url{http://arxiv.org/abs/1712.01815}.

\bibitem[{Silver et~al.(2017{\natexlab{b}})Silver, Schrittwieser, Simonyan,
  Antonoglou, Huang, Guez, Hubert, Baker, Lai, Bolton
  et~al.}]{silver2017mastering}
Silver D, Schrittwieser J, Simonyan K, Antonoglou I, Huang A, Guez A, Hubert T,
  Baker L, Lai M, Bolton A, et~al. (2017{\natexlab{b}}) Mastering the game of
  go without human knowledge. \emph{Nature} 550(7676):354--359.

\bibitem[{Sutton \protect\BIBand{} Barto(2018)}]{sutton2018reinforcement}
Sutton RS, Barto AG (2018) \emph{Reinforcement learning: An introduction} (MIT
  press).

\bibitem[{Tesauro \protect\BIBand{} Galperin(1996)}]{tesauro1996line}
Tesauro G, Galperin G (1996) On-line policy improvement using monte-carlo
  search. \emph{Advances in Neural Information Processing Systems} 9.

\bibitem[{Tomar et~al.(2020)Tomar, Efroni, \protect\BIBand{}
  Ghavamzadeh}]{tomar2020multistep}
Tomar M, Efroni Y, Ghavamzadeh M (2020) Multi-step greedy reinforcement
  learning algorithms. \emph{International Conference on Machine Learning},
  9504--9513 (PMLR).

\bibitem[{Uehara et~al.(2021)Uehara, Zhang, \protect\BIBand{}
  Sun}]{uehara2021representation}
Uehara M, Zhang X, Sun W (2021) Representation learning for online and offline
  rl in low-rank mdps. \emph{arXiv preprint arXiv:2110.04652} .

\bibitem[{Van Der~Wal(1978)}]{van1978discounted}
Van Der~Wal J (1978) Discounted markov games: Generalized policy iteration
  method. \emph{Journal of Optimization Theory and Applications}
  25(1):125--138.

\bibitem[{Winnicki et~al.(2021)Winnicki, Lubars, Livesay, \protect\BIBand{}
  Srikant}]{annaor}
Winnicki A, Lubars J, Livesay M, Srikant R (2021) The role of lookahead and
  approximate policy evaluation in policy iteration with linear value function
  approximation. \emph{CoRR} abs/2109.13419,
  \urlprefix\url{https://arxiv.org/abs/2109.13419}.

\bibitem[{Winnicki \protect\BIBand{} Srikant(2022)}]{annacdc}
Winnicki A, Srikant R (2022) Reinforcement learning with unbiased policy
  evaluation and linear function approximation. \emph{2022 IEEE 61st Conference
  on Decision and Control (CDC)}, 801--806,
  \urlprefix\url{http://dx.doi.org/10.1109/CDC51059.2022.9992427}.

\bibitem[{Winnicki \protect\BIBand{} Srikant(2023)}]{winnicki2023convergence}
Winnicki A, Srikant R (2023) On the convergence of policy iteration-based
  reinforcement learning with monte carlo policy evaluation. \emph{Artificial
  Intelligence and Statistics} .

\bibitem[{Xie et~al.(2020)Xie, Chen, Wang, \protect\BIBand{}
  Yang}]{xie2020learning}
Xie Q, Chen Y, Wang Z, Yang Z (2020) Learning zero-sum simultaneous-move markov
  games using function approximation and correlated equilibrium.
  \emph{Conference on learning theory}, 3674--3682 (PMLR).

\bibitem[{Yang et~al.(2020)Yang, Juntao, \protect\BIBand{}
  Lingling}]{yang2020multi}
Yang Y, Juntao L, Lingling P (2020) Multi-robot path planning based on a deep
  reinforcement learning dqn algorithm. \emph{CAAI Transactions on Intelligence
  Technology} 5(3):177--183.

\bibitem[{Yang \protect\BIBand{} Wang(2020)}]{yang2020overview}
Yang Y, Wang J (2020) An overview of multi-agent reinforcement learning from
  game theoretical perspective. \emph{arXiv preprint arXiv:2011.00583} .

\bibitem[{Zhang et~al.(2020)Zhang, Kakade, Basar, \protect\BIBand{}
  Yang}]{zhang2020model}
Zhang K, Kakade S, Basar T, Yang L (2020) Model-based multi-agent rl in
  zero-sum markov games with near-optimal sample complexity. \emph{Advances in
  Neural Information Processing Systems} 33:1166--1178.

\bibitem[{Zhang et~al.(2021)Zhang, Yang, \protect\BIBand{}
  Ba{\c{s}}ar}]{zhang2021multi}
Zhang K, Yang Z, Ba{\c{s}}ar T (2021) Multi-agent reinforcement learning: A
  selective overview of theories and algorithms. \emph{Handbook of
  reinforcement learning and control} 321--384.

\bibitem[{Zhang et~al.(2022)Zhang, Ren, Yang, Gonzalez, Schuurmans,
  \protect\BIBand{} Dai}]{zhang2022making}
Zhang T, Ren T, Yang M, Gonzalez J, Schuurmans D, Dai B (2022) Making linear
  mdps practical via contrastive representation learning. \emph{International
  Conference on Machine Learning}, 26447--26466 (PMLR).

\bibitem[{Zhang et~al.(2019)Zhang, Zhang, Miehling, \protect\BIBand{}
  Basar}]{zhang2019non}
Zhang X, Zhang K, Miehling E, Basar T (2019) Non-cooperative inverse
  reinforcement learning. \emph{Advances in neural information processing
  systems} 32.

\bibitem[{Zhao et~al.(2022)Zhao, Tian, Lee, \protect\BIBand{}
  Du}]{zhao2022provably}
Zhao Y, Tian Y, Lee J, Du S (2022) Provably efficient policy optimization for
  two-player zero-sum markov games. \emph{International Conference on
  Artificial Intelligence and Statistics}, 2736--2761 (PMLR).

\end{thebibliography}
\begin{APPENDICES}

\section{Proof of Theorem \ref{thm:theorem 1}}\label{appendix:appendix A}
The proof of Theorem \ref{thm:theorem 1} relies on showing a contraction property of the $T_{m,H}$ operator. 
\bigskip
We will show that $T_{m,H}$ is a contraction towards $J^*$ for all $m$ and $H$, including $m=\infty$ and all $H\geq 1.$ We note that this only holds for sufficiently large $H$ per Assumption \ref{assumption 1 games}. 

First, we have the following:
Since $T_{\mu_{k+1},\nu_{k+1}}V$ is defined as $g(\mu_{k+1},\nu_{k+1})+\alpha P(\mu_{k+1},\nu_{k+1})V,$ we can directly apply the contraction property from single player MDPs to see that:
\begin{align*}
\norm{T_{m,H}V_k-J^{\mu_{k+1},\nu_{k+1}}}_\infty &= \norm{T_{\mu_{k+1},\nu_{k+1}}^mT^{H-1}V_k - J^{\mu_{k+1},\nu_{k+1}}}_\infty \\&\leq \alpha^m \norm{T^{H-1}V_k - J^{\mu_{k+1},\nu_{k+1}}}_\infty.
\end{align*}
Thus, the following holds:
\begin{align}
\nonumber\norm{T_{m,H}V_k-T^{H-1}V_k}_\infty\nonumber \nonumber&=  \norm{T_{m,H}V_k- J^{\mu_{k+1},\nu_{k+1}}+J^{\mu_{k+1},\nu_{k+1}}  -T^{H-1}V_k}_\infty \\ \nonumber &\leq \norm{T_{m,H}V_k- J^{\mu_{k+1},\nu_{k+1}}}_\infty +\norm{J^{\mu_{k+1},\nu_{k+1}} -T^{H-1}V_k}_\infty  \\
 \nonumber &\leq \alpha^m \norm{T^{H-1}V_k - J^{\mu_{k+1},\nu_{k+1}}}_\infty \\&+\norm{T^{H-1}V_k - J^{\mu_{k+1},\nu_{k+1}}}_\infty \\
  &= (1+\alpha^m)\norm{T^{H-1}V_k - J^{\mu_{k+1},\nu_{k+1}}}_\infty.\label{eq:three}
\end{align}

We will now attempt to obtain a bound on $\norm{T^{H-1}V_k - J^{\mu_{k+1},\nu_{k+1}}}_\infty$. 
Here, we bypass a crucial monotonicity property of $T$ for single player systems that is not present in games. We  have by definition of $T_{\mu_{k+1},\nu_{k+1}}$ the following for all $\ell$:
\begin{align*}
&\norm{T_{\mu_{k+1},\nu_{k+1}}^{\ell+1}T^{H-1}V_k-T_{\mu_{k+1},\nu_{k+1}}^{\ell}T^{H-1}V_k}_\infty  \leq \alpha^\ell \norm{TV_k-V_k}_\infty.
\end{align*}
To start, we will need the following pseudo-contraction property of $T$ the optimal value function \cite{bertsekastsitsiklis}:
\begin{align*}
    \norm{TV-J^*}_\infty \leq \alpha\norm{V-J^*}_\infty.
\end{align*}
Since $T_{\mu_{k+1},\nu_{k+1}}T^{H-1}V_k = T^H V_k$ and using the property of $T$, we have the following:
\begin{align*}
    &\norm{T_{\mu_{k+1},\nu_{k+1}}T^{H-1}V_k - T^{H-1} V_k}_\infty\\ &=  \norm{T_{\mu_{k+1},\nu_{k+1}}T^{H-1}V_k -J^* +J^*- T^{H-1} V_k}_\infty\\
    &\leq \norm{T_{\mu_{k+1},\nu_{k+1}}T^{H-1}V_k -J^*}_\infty +\norm{J^*- T^{H-1} V_k}_\infty \\
    &= \norm{T^{H}V_k -J^*}_\infty +\norm{J^*- T^{H-1} V_k}_\infty \\
    &\leq \alpha^H \norm{V_k - J^*}_\infty + \alpha^{H-1}\norm{J^*-V_k}_\infty \\
    &= \underbrace{(\alpha^H + \alpha^{H-1})\norm{J^*-V_k}_\infty}_{=: \tilde{a}}.
\end{align*}

Thus,
\begin{align}
    -T_{\mu_{k+1},\nu_{k+1}}T^{H-1}V_k\leq  - T^{H-1} V_k + \tilde{a}.
\end{align}

Suppose that we apply the $T_{\mu_{k+1},\nu_{k+1}}$ operator $\ell-1$ times to both sides. Then, due to monotonicity and the fact $T_{\mu,\nu}(J+ce)=T_{\mu,\nu}(J)+\alpha ce,$ for any policy $(\mu,\nu),$ we have the following:
\begin{align*}
    {T^\ell_{\mu_{k+1},\nu_{k+1}}} T^{H-1}V_k \leq \alpha^{\ell } \tilde{a}e + {T^{\ell+1}_{\mu_{k+1},\nu_{k+1}}}T^{H-1}V_k.
\end{align*}
Using a telescoping sum, we get the following inequality:
\begin{align*}
    T_{\mu_{k+1},\nu_{k+1}}^j T^{H-1} V_k - T^{H-1}V_k
    &\geq - \sum_{\ell = 1}^{j} \alpha^{\ell - 1} \tilde{a} e.
\end{align*}
Taking the limit as $j\rightarrow\infty$ on both sides, we have the following:
\begin{align}
    J^{\mu_{k+1},\nu_{k+1}} - T^{H-1}V_k \geq - \frac{\tilde{a} e}{1-\alpha}.
\label{eq:one}
\end{align}
In the other direction, we have the following:
\begin{align}
     - T^{H-1} V_k\leq -T_{\mu_{k+1},\nu_{k+1}}T^{H-1}V_k + \tilde{a}.
\end{align}
Applying the $T_{\mu_{k+1},\nu_{k+1}}$ operator $\ell-1$ times to both sides, we have :
\begin{align*}
    {T^{\ell+1}_{\mu_{k+1},\nu_{k+1}}} T^{H-1}V_k \leq \alpha^{\ell } \tilde{a}e + {T^{\ell}_{\mu_{k+1},\nu_{k+1}}}T^{H-1}V_k.
\end{align*}
Using a telescoping sum, we get the following inequality:
\begin{align*}
   T^{H-1}V_k- T_{\mu_{k+1},\nu_{k+1}}^j T^{H-1} V_k 
    &\geq - \sum_{\ell = 1}^{j} \alpha^{\ell - 1} \tilde{a} e.
\end{align*}
Taking the limit as $j\rightarrow\infty$ on both sides, we have the following:
\begin{align}
   T^{H-1}V_k - J^{\mu_{k+1},\nu_{k+1}}  \geq - \frac{\tilde{ae} }{1-\alpha}.
   \label{eq:two}
   \end{align}
Hence, putting inequalities \eqref{eq:one} and \eqref{eq:two} together, we have the following bound:
\begin{align*}
    \norm{T^{H-1}V_k - J^{\mu_{k+1},\nu_{k+1}}}_\infty \leq \frac{\tilde{a} }{1-\alpha}.
\end{align*}

We plug this bound into our result in \eqref{eq:three} to get:
\begin{align}
  \nonumber&\nonumber\norm{T_{m,H}V_k-T^{H-1}V_k}_\infty\nonumber \nonumber\\&\nonumber\leq \frac{\tilde{a} e}{1-\alpha} \\&=  \frac{(1+\alpha^m)(\alpha^H + \alpha^{H-1})\norm{J^*-V_k}_\infty }{1-\alpha}.
\label{eq:four} \end{align}

We now provide the reverse triangle inequality which we will use in the next step: \begin{align*}
    \norm{X-Y}_\infty - \norm{Y-Z}_\infty \leq \norm{X-Z}_\infty \forall X,Y,Z.
\end{align*} Using the reverse triangle inequality we have the following bound:
\begin{align*}
&\norm{T_{m,H}V_k - J^*}_\infty - \norm{T^{H-1}V_k-J^*}\\&\leq \norm{T_{m,H}V_k-T^{H-1}V_k}_\infty.
\end{align*}

Now, we use the pseudo-contraction property of $T$ towards $J^*$ as follows:
\begin{align*} &\norm{T_{m,H}V_k - J^*}_\infty \\&\leq \norm{T_{m,H}V_k-T^{H-1}V_k}_\infty +  \norm{J^*-T^{H-1}V_k}_\infty\\
&\leq \norm{T_{m,H}V_k-T^{H-1}V_k}_\infty + \alpha^{H-1} \norm{J^*-V_k}_\infty \\
&\leq \frac{(1+\alpha^m)(\alpha^H + \alpha^{H-1})\norm{J^*-V_k}_\infty }{1-\alpha}+\alpha^{H-1} \norm{J^*-V_k}_\infty, 
\end{align*}
where in the last line we plug in our bound in \eqref{eq:four}.

Hence, the following holds:

\begin{align*}
  &\norm{T_{m,H}V_k-J^*}_\infty  \\&\leq  \Big(\alpha^{H-1}+(1+\alpha^m )\frac{\alpha^{H-1}}{1-\alpha}(1+\alpha)\Big)\norm{V_k - J^*}_\infty.
\end{align*}

Noting that $V_{k+1} = T_{m,H}V_k,$ we iterate to get the following bound:
\begin{align*}
&\norm{V_k-J^*}_\infty \\& \leq  \Big(\alpha^{H-1}+(1+\alpha^m )\frac{\alpha^{H-1}}{1-\alpha}(1+\alpha)\Big)^k\norm{V_0 - J^*}_\infty,
\end{align*} and, further using Assumption \ref{assumption 1 games}, we have that $V_k \to J^*,$ which proves the exponential rate of convergence of our algorithm.
endproof

The framework of the techniques we use are based on those of the work of \cite{winnicki2023convergence} however unlike the work of \cite{winnicki2023convergence} the setting of the problem is a two-player game in a deterministic setting as opposed to online learning with stochastic approximation for a single-player system.

\hfill
\Halmos
\section{Proof of Theorem \ref{theorem 1 MG}} \label{appendix: appendix B}
First, we define $V_k$ as follows:
\begin{align*}
    V_k := \Phi \theta_k.
\end{align*}
In order to prove Theorem \ref{alg:alg 4 MG}, we will first give Lemma \ref{lemma 1 MG}. Lemma \ref{lemma 1 MG} is proved in \cite{winnicki2023convergence} for the MDP setting and can easily be extended to the zero-sum Markov games setting using contraction properties given in Section \ref{prelim}.
\begin{lemma}
\begin{align*}
\norm{J^{{\mu_{k+1},\nu_{k+1}}}-T^{H-1}V_k}_\infty \leq \frac{\alpha^{H-1}}{1-\alpha}\norm{TV_k - V_k}_\infty,
\end{align*} \label{lemma 1 MG}
\end{lemma} where $J^{{\mu_{k+1},\nu_{k+1}}}$ is the value function corresponding to policy ${\mu_{k+1},\nu_{k+1}}$. 

We note that Lemma \ref{lemma 1 MG} implies:
\begin{align*}
    &\norm{J^{{\mu_{k+1},\nu_{k+1}}}-T^{H-1}V_k}_\infty \\&\leq \frac{\alpha^{H-1}}{1-\alpha}\norm{TV_k - V_k}_\infty \allowdisplaybreaks\\
    &=\frac{\alpha^{H-1}}{1-\alpha}\norm{TV_k-J^*+J^* - V_k}_\infty\allowdisplaybreaks\\
    &\leq \frac{\alpha^{H-1}}{1-\alpha}\norm{TV_k-J^*}_\infty+\frac{\alpha^{H-1}}{1-\alpha}\norm{J^* - V_k}_\infty\allowdisplaybreaks\\
    &\leq \frac{\alpha^{H}}{1-\alpha}\norm{V_k-J^*}_\infty+\frac{\alpha^{H-1}}{1-\alpha}\norm{J^* - V_k}_\infty\allowdisplaybreaks\\
    &= \Big(\frac{\alpha^{H-1}(1+\alpha)}{1-\alpha}\Big)\norm{V_k-J^*}_\infty.
\end{align*}
Note that the above implies the following:
\begin{align*}
   &T^{H-1}\Phi \theta_k - \Big(\frac{\alpha^{H-1}(1+\alpha)}{1-\alpha}\Big)\norm{V_k-J^*}_\infty 
   \\&\leq J^{{\mu_{k+1},\nu_{k+1}}} 
   \\& \leq T^{H-1}\Phi \theta_k +\Big(\frac{\alpha^{H-1}(1+\alpha)}{1-\alpha}\Big)\norm{V_k-J^*}_\infty.
\end{align*}
We will now upper bound our iterates $V_{k+1}$:
\begin{align}
V_{k+1} \nonumber&=\Phi \theta_{k+1}
\nonumber\\&= \scriptM T_{{\mu_{k+1},\nu_{k+1}}}^m T^{H-1}V_k \nonumber\\
\nonumber&= \scriptM T_{{\mu_{k+1},\nu_{k+1}}}^m T^{H-1}V_k-J^{{\mu_{k+1},\nu_{k+1}}} \allowdisplaybreaks\\&+ J^{{\mu_{k+1},\nu_{k+1}}}\nonumber\\
\nonumber&\leq \norm{\scriptM T_{{\mu_{k+1},\nu_{k+1}}}^m T^{H-1}V_k-J^{{\mu_{k+1},\nu_{k+1}}}}_\infty\\\nonumber    & + J^{{\mu_{k+1},\nu_{k+1}}}\nonumber\allowdisplaybreaks\\
\nonumber&\leq \norm{\scriptM T_{{\mu_{k+1},\nu_{k+1}}}^m T^{H-1}V_k-\scriptM J^{{\mu_{k+1},\nu_{k+1}}}}_\infty\nonumber\\&+\norm{\scriptM J^{{\mu_{k+1},\nu_{k+1}}} - J^{{\mu_{k+1},\nu_{k+1}}}}_\infty \allowdisplaybreaks\nonumber\\&+ J^{{\mu_{k+1},\nu_{k+1}}}\nonumber\allowdisplaybreaks\\
\nonumber&\leq \delta_{FV} \alpha^m \norm{ T^{H-1}V_k- J^{{\mu_{k+1},\nu_{k+1}}}}_\infty+\delta_{app} \nonumber\\&+ J^{{\mu_{k+1},\nu_{k+1}}}\nonumber\allowdisplaybreaks\\
\nonumber&\leq  T^{H-1}\Phi \theta_k + \delta_{app} \\&+\frac{(\delta_{FV} \alpha^{m+H-1}+\alpha^{H-1})(1+\alpha) }{1-\alpha}\Big)\norm{V_k-J^*}_\infty, \label{eq: before lemma 2 unbiased full}
\end{align} 
where $\delta_{FV} :=\norm{\scriptM}_\infty$ and  $\delta_{app} := \sup_{k,\mu_k}\norm{\scriptM J^{\mu}-J^{\mu}}_\infty$.

We can follow the above steps to derive an analogous lower bound as follows:
\begin{align}
V_{k+1} &=\Phi \theta_{k+1}
\nonumber\\&= \scriptM T_{{\mu_{k+1},\nu_{k+1}}}^m T^{H-1}V_k \nonumber\allowdisplaybreaks\\
\nonumber&= \scriptM T_{{\mu_{k+1},\nu_{k+1}}}^m T^{H-1}V_k-J^{{\mu_{k+1},\nu_{k+1}}} + J^{{\mu_{k+1},\nu_{k+1}}}\nonumber\allowdisplaybreaks\\
\nonumber&\geq -\norm{\scriptM T_{{\mu_{k+1},\nu_{k+1}}}^m T^{H-1}V_k-J^{{\mu_{k+1},\nu_{k+1}}}}_\infty \nonumber\\&+ J^{{\mu_{k+1},\nu_{k+1}}}\nonumber\allowdisplaybreaks\\
\nonumber&\geq -\norm{\scriptM T_{{\mu_{k+1},\nu_{k+1}}}^m T^{H-1}V_k-\scriptM J^{{\mu_{k+1},\nu_{k+1}}}}_\infty\nonumber\\&+\norm{\scriptM J^{{\mu_{k+1},\nu_{k+1}}} - J^{{\mu_{k+1},\nu_{k+1}}}}_\infty \nonumber\\&+ J^{{\mu_{k+1},\nu_{k+1}}}\nonumber\allowdisplaybreaks\\
\nonumber&\geq -\delta_{FV} \alpha^m \norm{ T^{H-1}V_k- J^{{\mu_{k+1},\nu_{k+1}}}}_\infty-\delta_{app} \allowdisplaybreaks\\&+ J^{{\mu_{k+1},\nu_{k+1}}}\nonumber\allowdisplaybreaks\\
\nonumber&\geq  T^{H-1}\Phi \theta_k -\delta_{app} \\&-\frac{(\delta_{FV} \alpha^{m+H-1}+\alpha^{H-1})(1+\alpha) }{1-\alpha}\Big)\norm{V_k-J^*}_\infty\allowdisplaybreaks, \label{eq: before lemma 2 unbiased full 2}
\end{align} 
We now use the above to obtain the following bound on $\norm{V_{k+1}-T^{H-1}\Phi \theta_k}_\infty:$
\begin{align*}
    &\norm{V_{k+1}-T^{H-1}\Phi \theta_k}_\infty\\&\leq \delta_{app} +\frac{(\delta_{FV} \alpha^{m+H-1}+\alpha^{H-1})(1+\alpha) }{1-\alpha}\norm{V_k-J^*}_\infty.
\end{align*}
Using the reverse triangle inequality, we have:
\begin{align*}
&\norm{V_{k+1}-T^{H-1}\Phi \theta_k}_\infty \\
&=\norm{V_{k+1}-J^*+J^*-T^{H-1}\Phi \theta_k}_\infty \\
&\geq \norm{V_{k+1}-J^*}_\infty-\norm{J^*-T^{H-1}\Phi \theta_k}_\infty,
\end{align*}
which implies that
\begin{align*}
    &\norm{V_{k+1}-J^*}_\infty \\&\leq \norm{T^{H-1}V_k-J^*}_\infty + \delta_{app} \\&+\frac{(\delta_{FV} \alpha^{m+H-1}+\alpha^{H-1})(1+\alpha) }{1-\alpha}\norm{V_k-J^*}_\infty \\
    &\leq \Big(\alpha^{H-1}+\frac{(\delta_{FV} \alpha^{m+H-1}+\alpha^{H-1})(1+\alpha) }{1-\alpha}\Big) \norm{V_k-J^*}_\infty \\&+ \delta_{app}.
\end{align*}

Iterating over $k$, we have:
\begin{align*}
&\norm{V_{k}-J^*}_\infty \\&\leq \underbrace{\Big(\alpha^{H-1}+\frac{(\delta_{FV} \alpha^{m+H-1}+\alpha^{H-1})(1+\alpha) }{1-\alpha}\Big)^k \norm{V_0-J^*}_\infty }_{finite-time \text{ } component}\\&+\underbrace{\frac{\delta_{app}}{ 1-\Big(\alpha^{H-1}+\frac{(\delta_{FV} \alpha^{m+H-1}+\alpha^{H-1})(1+\alpha) }{1-\alpha}\Big)}}_{asymptotic \text{ }component}.
\end{align*}
\hfill
\Halmos
\section{Proof Of Theorem \ref{thm:theorem 3 unbiased full}} \label{appendix: appendix C}
Setting $$H(V_k) := E[\scriptM_k( T_{{\mu_{k+1},\nu_{k+1}}}^m T^{H-1}V_k+w_k)-J^{{\mu_{k+1},\nu_{k+1}}}|\scriptF_k],$$ we have the following bound:
\begin{align*}
    &\norm{H(V_k)-J^{{\mu_{k+1},\nu_{k+1}}}}_\infty \allowdisplaybreaks\\
    &= ||E[\scriptM_k( T_{{\mu_{k+1},\nu_{k+1}}}^m T^{H-1}V_k+w_k)-J^{{\mu_{k+1},\nu_{k+1}}}|\scriptF_k]||_\infty \allowdisplaybreaks\\
    &\leq ||E[\scriptM_k( T_{{\mu_{k+1},\nu_{k+1}}}^m T^{H-1}V_k+w_k)\\&-\scriptM_k( J^{{\mu_{k+1},\nu_{k+1}}}+ w_k)|\scriptF_k]||_\infty\\&+\norm{E[\scriptM_k( J^{{\mu_{k+1},\nu_{k+1}}}+ w_k)-J^{{\mu_{k+1},\nu_{k+1}}}|\scriptF_k]}_\infty \allowdisplaybreaks\\
    &\leq ||E[\scriptM_k( T_{{\mu_{k+1},\nu_{k+1}}}^m T^{H-1}V_k+w_k)\\&-\scriptM_k( J^{{\mu_{k+1},\nu_{k+1}}}+ w_k)|\scriptF_k]||_\infty\\&+\underbrace{\sup_{k, \mu_k}\norm{E[\scriptM_k( J^{{\mu_{k+1},\nu_{k+1}}}+ w_k)-J^{{\mu_{k+1},\nu_{k+1}}}|\scriptF_k]}_\infty}_{=: \delta'_{app}} \allowdisplaybreaks\\ 
    &= ||E[\scriptM_k( T_{{\mu_{k+1},\nu_{k+1}}}^m T^{H-1}V_k)-\scriptM_k( J^{{\mu_{k+1},\nu_{k+1}}})|\scriptF_k]||_\infty\\&+ \delta'_{app} \allowdisplaybreaks\\ 
    &= E[||\scriptM_k( T_{{\mu_{k+1},\nu_{k+1}}}^m T^{H-1}V_k)-\scriptM_k( J^{{\mu_{k+1},\nu_{k+1}}})||_\infty|\scriptF_k]\\&+ \delta'_{app} \allowdisplaybreaks\\ 
    &\leq  E[\sup_k \norm{\scriptM_k}_\infty ||T_{{\mu_{k+1},\nu_{k+1}}}^m T^{H-1}V_k- J^{{\mu_{k+1},\nu_{k+1}}}||_\infty|\scriptF_k]\\&+ \delta'_{app} \allowdisplaybreaks\\ 
    &=  \underbrace{\sup_k \norm{\scriptM_k}_\infty}_{=: \delta'_{FV}} || T_{{\mu_{k+1},\nu_{k+1}}}^m T^{H-1}V_k -J^{{\mu_{k+1},\nu_{k+1}}}||_\infty\allowdisplaybreaks\\&+ \delta'_{app} \allowdisplaybreaks\\ 
    &\leq \alpha^m \delta'_{FV}|| T^{H-1}V_k -J^{{\mu_{k+1},\nu_{k+1}}}||_\infty+ \delta'_{app}.\allowdisplaybreaks
\end{align*}
Adding and subtracting $T^{H-1}V_k$ and using the triangle inequality, we furthermore have:
\begin{align}
   \nonumber &\norm{H(V_k)-T^{H-1}V_k}_\infty \\
    \nonumber&\leq  (1+ \alpha^m \delta'_{FV}) \norm{  T^{H-1}V_k -J^{{\mu_{k+1},\nu_{k+1}}}}_\infty+ \delta'_{app} \\
    &\leq \frac{(1+ \alpha^m \delta'_{FV})\alpha^{H-1}}{1-\alpha} \norm{TV_k - V_k}_\infty+ \delta'_{app}, \label{eq: label ineq MG} 
\end{align} where the last line follows from Lemma \ref{lemma 1 MG}. Adding and subtracting $J^*$ on both sides of the inequality in \eqref{eq: label ineq MG} and using the triangle inequality, we have:
\begin{align}
   \nonumber &\norm{H(V_k)-J^*}_\infty 
    \\ \nonumber &\leq (\alpha^{H-1}+(1+\alpha)\frac{(1+ \alpha^m \delta'_{FV})\alpha^{H-1}}{1-\alpha})\norm{V_k - J^*}_\infty \\&+ \delta'_{app}. \label{eq: final eq MG FA}
\end{align}

We now provide the following Lemma \ref{lemma 2 MG FA} from Appendix D of \cite{winnicki2023convergence} which will be used to obtain Theorem \ref{thm:theorem 3 unbiased full}.
\begin{lemma} \label{lemma 2 MG FA}
    Suppose that we have the following sequence:
    \begin{align*}
    V_{k+1} = (1-\gamma_k) V_k + \gamma_k (H(V_k) + z_k),
\end{align*}
where 
\begin{align*}
    \norm{H(V_k)-J^*}_\infty \leq \beta\norm{V_k-J^*}_\infty + \delta
\end{align*} for some $0<\beta<1$, $\delta$ and $J^*.$ Then, the following holds: 
\begin{align*}
   \limsup_{k\to \infty} \norm{V_k - J^*}_\infty \leq  \frac{\delta}{1-\beta}.
\end{align*} \hfill $\diamond$
\end{lemma}

Applying Lemma \ref{lemma  2 MG FA} to \eqref{eq: final eq MG FA} with our $\beta$ in Lemma \ref{lemma 2 MG FA} being $\alpha^{H-1}+(1+\alpha)\frac{(1+ \alpha^m \delta'_{FV})\alpha^{H-1}}{1-\alpha}$ and $\delta$ of Lemma \ref{lemma 2 MG FA} being $\delta'_{app},$ we obtain Theorem \ref{thm:theorem 3 unbiased full} when Assumption \ref{assumption 2 MG} is satisfied.  
\hfill
\Halmos
\section{Proof of Theorem \ref{thm:theorem 3}} \label{appendix:appendix D}
We note that the settings of the bound in \cite{zhang2020model} are the same as the settings of our work, so we can directly apply the bounds in \cite{zhang2020model}. We will restate Theorem 3.3 from \cite{zhang2020model} in Lemma \ref{lemma 1 games}:

\begin{lemma}\label{lemma 1 games}
    Consider any $\epsilon, \delta>0$ with $\epsilon \in (0, 1/(1-\alpha)^{1/2}]$ and $\delta \in [0,1]$. When the number of samples of each state-actions tuple is at least $N$ by the learning oracle and a planning oracle is used based on the empirical model $\hat{\mathcal{G}}$ which is reward-aware and the entries are of the probability transition matrix is constructed by taking averages determined by the learning oracle to determine a policy $(\hat{\mu},\hat{\nu})$ where:
    \begin{align*}
        \norm{\hat{V}^{\hat{\mu},\hat{\nu}}-\hat{J}^*}_\infty \leq\epsilon_{opt},
    \end{align*} where $\hat{J}^*$ is the optimal value function for model $\hat{\mathcal{G}}$, when 
\begin{align*}
 &N \geq \frac{c \alpha \log\big[ c |\scriptS||\scriptU||\scriptV|(1-\alpha)^{-2}\delta^{-1}\big]}{(1-\alpha)^3 \epsilon^2} 
\end{align*} where for some absolute constant $c$, it holds that with probability at least $1-\delta,$ 
\begin{align*}
    \norm{Q^{{\hat{\mu},\hat{\nu}}}-Q^*}_\infty &\leq \frac{2\epsilon}{3}+\frac{5 \alpha \epsilon_{opt}}{1-\alpha}, \\ \norm{\hat{Q}^{\hat{\mu},\hat{\nu}}-Q^*}_\infty &\leq \epsilon + \frac{9\alpha \epsilon_{opt}}{1-\alpha}.
\end{align*}
\hfill $\diamond$
\end{lemma}

Now, we compute the complexity as follows:
It is easy to see that to compute $$Q(s, u, v) \leftarrow r(s, u, v)+ \alpha \sum_{s' \in \scriptD_\scriptR} P(s'|s, u,v)V(s') \forall (s,u,v) \in \scriptD$$ requires $d[2r+1]$ computations, computing $$\theta \leftarrow \displaystyle\argmin_\theta  \sum_{(s, u,v)\in \scriptD} [\phi(s,u,v)^\top \theta - Q(s,u,v)]^2$$ requires $d^3/3$ computations (as an upper bound), computing $$V(s') \leftarrow \displaystyle\max_{\substack{\mu\in \mathbb{R}^{|\scriptU(s')|} \\ \sum \mu_i=1\\0\leq \mu_i \leq 1  }} \displaystyle\min_{\substack{\nu(s) \in \mathbb{R}^{|\scriptV(s')|} \\ \sum \nu_i=1\\0\leq \nu_i \leq 1  }} \mu^\top A_{\theta, s'} \nu,$$ needs $r\times|\scriptA|_{max}^2\times d$ computations noting that there is a turn-based MDP which is well known to involve only deterministic policies, and since the $\mu^\top A_{\theta, s'} \nu$ is a special case of the previous step, no more additional computations are needed to determine an upper bound. Using the above, some algebra gives Theorem \ref{thm:theorem 3}.

\hfill
\hfill
\Halmos

\end{APPENDICES}

\end{document}